\documentclass[runningheads]{llncs}

\usepackage[utf8]{inputenc}

\usepackage{xcolor}
\usepackage{amsmath} 
\usepackage{amsfonts}
\usepackage{graphicx}
\usepackage{subfigure}

\makeatletter
\newcommand{\printfnsymbol}[1]{%
  \textsuperscript{\@fnsymbol{#1}}%
}
\makeatother

\begin{document}
\title{3D dynamic hand gestures recognition using the Leap Motion sensor and convolutional neural networks}
\titlerunning{3D dynamic hand gestures recognition}
%
\author{Katia Lupinetti\inst{1}\orcidID{0000-0002-0202-4909}\thanks{These authors contributed equally to this work.} \and
	Andrea Ranieri\inst{1}\orcidID{0000-0001-8317-3158}\printfnsymbol{1} \and
	Franca Giannini\inst{1}\orcidID{0000-0002-3608-6737} \and
	Marina Monti\inst{1}\orcidID{0000-0001-5198-7798}}

\authorrunning{K. Lupinetti et al.}
%
\institute{Istituto di Matematica Applicata e Tecnologie  Informatiche ``Enrico Magenes", CNR Via De Marini 6, 16149 Genova, Italy\\
	\email{katia.lupinetti@cnr.it},
	\email{andrea.ranieri@cnr.it},
	\email{franca.giannini@cnr.it},
	\email{marina.monti@cnr.it}
}

\maketitle              
\begin{abstract}
	Defining methods for the automatic understanding of gestures is of paramount importance in many application contexts and in Virtual Reality applications for creating more natural and easy-to-use human-computer interaction methods. In this paper, we present a method for the recognition of a set of non-static gestures acquired through the Leap Motion sensor. The acquired gesture information is converted in color images, where the variation  of hand joint positions during the gesture are projected on a plane and temporal information is represented with color intensity of the projected points. The classification of the gestures is performed using a deep Convolutional Neural Network (CNN). A modified version of the popular ResNet-50 architecture is adopted, obtained by removing the last fully connected layer and adding a new layer with as many neurons as the considered gesture classes. The method has been successfully applied to the existing reference dataset and preliminary tests have already been performed for the real-time recognition of dynamic gestures performed by users.
	
\end{abstract}

\section{Introduction}
Gesture recognition is an interesting and active research area whose applications are numerous and various, including, for instance, robotics, training systems, virtual prototyping, video surveillance, physical rehabilitation, and computer games. 
This wide interest is due to the fact that hands and fingers are used to communicate and to interact with the physical world \cite{ahmad2019hand}; then, by analyzing human gestures it is possible to improve the understanding of the non-verbal human interaction. This understanding poses the basis for the creation of more natural human-computer interaction, which is fundamental for the creation of immersive virtual environments with a high sense of presence.
Despite this popularity and interest, until few years ago, finger movements were difficult to be acquired and characterized, especially without the use of sophisticated tracking devices, which usually turn to be quite unnatural.
Indeed, there exist many methods trying to solve hand gesture recognition by using wearable devices \cite{dipietro2008survey,bourke2007evaluation,kevin2004trajectory}.
With the recent technology improvement, fingers' tracks can be digitally obtained relying only on RGB cameras eventually enhanced with depth information.
In this manner, it is possible abstracting human hands by adopting two main representations: 3D model-based and appearance-based \cite{rautaray2015vision}. Generally, 3D model-based representations are deduced by exploiting depth information, but there are methods trying to reconstruct 3D hand representations using only RGB-data. 

Hand gestures can be classified as \textit{static}, i.e. if no change occurs over time, or \textit{dynamic}, i.e. if several hand poses contribute to the final semantic of the gesture within an arbitrary time interval. So far, several works address static gesture, focusing on pre-defined gesture vocabularies, such as the recognition of the sign language of different countries \cite{kumar20203d,ravi2019multi,kaluri2018optimized,kumar2018independent,mapari2016american,kuznetsova2013real}.
Even if dynamic gestures are not universal but vary in different countries and cultures, they are more natural and intuitive than the static ones.

Since a tracking without gloves or controllers is more natural and efficient for users, in this paper we aim at defining a method for dynamic gesture recognition based on 3D hand representation reconstructed from the Leap Motion sensor tracking.
Our method relies on deep learning techniques applied to the images obtained by plotting the positions of the hands acquired over time on a specific 2D plane, condensing the temporal information uniquely as traces left by the fingertips that fade towards a value of transparency (the alpha value) equal to zero as time passes. Compared to also drawing the traces of the other edges that make up the hand, we have found that this approach maximizes the information that can be condensed into a single image while keeping it understandable for humans.

For the training, a public available dataset presenting 1134 gestures \cite{boulahia2017dynamic,LMDHG_dataset} has been used. The first stage of the evaluation of the deep neural network has been carried out on a subset of the 30\% of the available gestures, mantaining the split as presented in the original paper \cite{boulahia2017dynamic} and reaching an overall accuracy of the 91.83\%. We also propose our own dataset with about 2000 new dynamic gesture samples, created following considerations on the balance, number of samples and noise of the original dataset. We will show that by using our approach and our dataset, it is possible to exceed 98\% accuracy in the recognition of dynamic hand gestures acquired through the Leap Motion sensor. Finally, we will briefly talk about the real-time setup and how it has already been successfully used to acquire the new dataset proposed in this paper and to perform some preliminary user tests.


%

\clearpage	
The rest of the paper is organized as follows.
Section \ref{RelatedWorks} reviews the most pertinent related works. Section \ref{Method} and \ref{Experiments} detail the proposed method and show the results of the experimentation carried out respectively.
Finally, section \ref{Conclusions} ends the paper providing conclusions and future steps.

\section{Related works}\label{RelatedWorks}
The ability of recognizing hand gestures, or more in general understanding the interaction between humans and the surrounding environment, has arisen interests in numerous fields and has been tackled in several studies consequently.
So far, several commercial sensors for capturing full hand and finger action are available on the market, generally they can be divided into \textit{wearable} (such as data gloves) and \textit{external} devices (such as video cameras). Wearable sensors can address different purposes, for instance VR Glove by Manus\footnote{https://manus-vr.com/}, Cyber Glove System\footnote{http://www.cyberglovesystems.com/}, Noitom Hi5 VR\footnote{https://hi5vrglove.com/} are designed mainly for VR training; while the Myo Gesture Control Armband is especially used in medical applications \cite{bachmann2018review}.
This kind of technology is very accurate and with fast reaction speed. 
However, using gloves requires a calibration phase every time a different user starts and not always allows natural hand gestures and intuitive interaction because the device itself could constrain fingers motion \cite{abraham2018hand,gunawardane2017comparison,lawson2016future,sharp2015accurate}.
Therefore, research on hand motion tracking has begun investigating vision-based techniques relying on external devices with the purpose of allowing a natural and direct interaction \cite{ahmad2019hand}.

In the following sections, we review methods registering hands from RGB cameras (both monocular or stereo) and RGB-D cameras and interacting through markerless visual observations.

\subsection{Methods based on RGB sensors}
The use of simple RGB cameras for the hand tracking, and consequently for their gesture recognition, is a challenging problem in computer vision. 
So far, works using markerless RGB-images mainly aim at the simple tracking of the motion, as the body movement \cite{khokhlova20183d,mehta2017vnect,bobick2001recognition} or the hand skeleton \cite{GANHands2018,romero2010hands,stenger2006model}; while the motion recognition and interpretation has still big room for improvement.
Considering the method proposed in \cite{GANHands2018}, which presents an approach for real-time hand tracking from monocular RGB-images, it allows the reconstruction of the 3D hand skeleton even if occlusions occur. As a matter of principle, this methodology could be used as input for a future gesture recognition. Anyhow, it outperforms the RGB-methods but not the RGB-D ones presenting some difficulties when the background has similar appearance as the hand and when multiple hands are close in the input image.

Focusing on hand gesture recognition, Barros et al. \cite{barros2014real} propose a deep neural model to recognize dynamic gestures with minimal image pre-processing and real time recognition. Despite the encouraging results obtained by the authors, the recognized gestures are significantly different from each other, so the classes are well divided, which usually greatly simplifies the recognition of the gestures.

Recently, \cite{santos2019dynamic} proposes a system for the 3D dynamic hand gesture recognition by a deep learning architecture that uses a Convolutional Neural Network (CNN) applied on Discrete Fourier Transform on the artificial images. 
The main limitation of this approach is represented by the acquisition setup, i.e. it must be used in an environment where the cameras are static or where the relative movement between the background and the person is minimal.



\subsection{Methods based on Depth sensors}
To avoid many issues related to the use of simple RGB-images, depth cameras are widely used for hand tracking and gesture recognition purposes. Generally, the most common used depth cameras are the Microsoft Kinect\footnote{https://developer.microsoft.com/en-us/windows/kinect/} and the  Leap Motion (LM) sensor\footnote{https://developer.leapmotion.com}.\\
The Kinect sensor includes a QVGA (320x240) depth camera and a VGA (640x480) video camera, both of which produce image streams at 30 frames per seconds (fps). The sensor is limited by near and far thresholds for depth estimation and it is able to track the full-body \cite{suarez2012hand}.
The LM is a compact sensor that exploits two CMOS cameras capturing images with a frame rate of 50 up to 200fps \cite{ameur2016comprehensive}. It is very suitable for hand gesture recognition because it is explicitly targeted to hand and finger tracking.
Another type of sensor that is adopted sometimes is the Time-of-Flight camera, which measures  distance between the camera and the subject for each point of the image by using an artificial light signal provided by a laser or an LED. This type of sensor has a low resolution (176x144) and it is generally paired with a higher resolution RGB camera \cite{van2011combining}.

Using one of the above mentioned sensors, there are several works that address the recognition of \textit{static} hand gestures. 
Mapari and Kharat \cite{mapari2016american} proposed a method to recognize the American Sign Language (ASL). Using the data extracted from the LMC, they compute 48 features (18 positional values, 15 distance values and 15 angle values) for 4672 collected signs (146 users for 32 signs) feeding an artificial neural network by using a Multilayer Perceptron (MLP).
Filho et al. \cite{stinghen2016gesture} use the normalized positions of the five finger tips and the four angles between adjacent fingers as features for different classifiers (K-Nearest Neighbors, Support Vector Machines and Decision Trees). They compare the effectiveness of the proposed classifiers over a dataset of 1200 samples (6 uses for 10 gestures) discovering that the Decision Trees is the method that better performs.
Still among the methods to recognize static postures, Kumar et al. \cite{kumar2018independent} apply an Independent Bayesian Classification Combination (IBCC) approach.
Their idea is to combine hand features extracted by LM (3D fingertip positions and 3D palm center) with face features acquired by Kinect sensor (71 facial 3D points) in order to improve the meaning associated with a certain movement. One challenge performing this combination relies on fusion of the features, indeed pre-processing techniques are necessary to synchronize the frames since the two devices are not comparable. 

A more challenging task, which increases the engagement by a more natural and intuitive interaction, is the recognition of \textit{dynamic} gestures. In this case, it is crucial preserving spatial and temporal information associated with the user movement.
Ameur et al. \cite{ameur2016comprehensive} present an approach for the dynamic hand gesture recognition extracting spatial features through the 3D data provided by a Leap Motion sensor and feeding a Support Vector Machine (SVM) classifier based on the one-against-one approach.
With the aim of exploiting also the temporal information, Gatto et al. \cite{gatto2017orthogonal} propose a representation for hand gestures exploiting the Hankel matrix to combine gesture images generating a sub-space that preserves the time information. Then, gestures are recognized supposing that if the distance between two sub-spaces is small enough, then these sub-spaces are similar to each other.
Mathe et al. \cite{mathe2018arm} create artificial images that encode the movement in the 3D spaces of skeletal joints tracked by a Kinect sensor. Then, a deep learning architecture that uses a CNN is applied on the Discrete Fourier Transformation of the artificial images. With this work, authors demonstrate that is possible to recognize hand gestures without the need of a feature extraction phase.
Boulahia et al. \cite{boulahia2017dynamic} extract features on the hands trajectories, which describe \textit{local information}, for instance describing the starting and ending 3D coordinates of the 3D pattern resulting from trajectories assembling, and \textit{global information}, such as the convex hull based feature. Temporal information is considered by extracting features on overlapping sub-sequences resulting from a temporal split of the global gesture sequence. In this way, the authors collect a vector of 356 elements used to feed a SVM classifier.

In general, the use of complex and sophisticated techniques to extract ad-hoc features and manage temporal information requires more human intervention and does not scale well when the dictionary of gestures to be classified has to be expanded. Furthermore, the extraction of hundreds of features at different time scales may even take more CPU time than a single forward pass on a standard CNN already optimized against modern GPU architectures, thus not guaranteeing real-time performance in classification.


\section{Overview of the proposed approach}\label{Method}
Based on the assumption that natural human-computer interaction should be able to recognize not only predefined postures but also dynamic gestures, here we propose a method for the automatic recognition of gestures using images obtained from LM data. Our method uses state-of-the-art deep learning techniques, both in terms of the CNN architectures and the training and gradient descent methods employed.


In the following sub-sections, first we describe the problem of dynamic gesture recognition from images (Section \ref{problemFormulation}), then we illustrate the pipeline to create the required images and how we feed them to the neural network model (Section \ref{pipeline}). Finally, we introduce the LMDHG dataset adopted and the rationale that led us to use it (Section \ref{DatasetDescription}).

\subsection{Problem formulation}\label{problemFormulation}
Let $g_{i}$ be a dynamic gesture and $S = \{C_{h}\}_{h=1}^{N}$ a set of gesture classes, where $N$ identifies the number of classified gestures.
The variation of $g_{i}$ over time can be defined as:
\begin{equation}
G_{i} = \Big\{\mathcal{G}_{i}^{\tau} \Big\}_{\tau = 1}^{T_{i}},
\end{equation}
where $\tau \in [1, T_{i}]$ defines a certain instant in a temporal window of size $T_{i}$ and $\mathcal{G}_{i}^{\tau}$ represents the frame of $g_{i}$ at the time $\tau$. 
Note that a gesture can be performed over a variable temporal window (depending on the gesture itself or on the user aptitude).
The dynamic hand gesture classification problem can be defined as finding the class $C_{h}$ where $g_{i}$ \textit{most likely} belongs to, i.e. finding the pair $(g_{i}, C_{h})$ whose probability distribution $\mathbb{P}(g_{i}, C_{h})$ has the maximum value $\forall h$.

Let $\Phi$ be a mapping that transforms the space and the temporal information associated with a gesture $g_{i}$ resulting into a single image defined as:
\begin{equation}
I_{i} = \Phi(G_{i}).
\end{equation}
With this representation, there exist a single $I_{i}$ for each gesture $g_{i}$ regardless of the temporal window size $T_{i}$.
This new representation encodes in a more compact manner the different instants $\tau$ of each gesture and represents the new data to be recognized and classified. 
Then, the classification task can be redefined in finding whether an image $I_{i}$ belongs to a certain gesture class $C_{h}$, i.e. finding the pair $(I_{i}, C_{h})$ whose probability distribution $\mathbb{P}(I_{i}, C_{h})$ has the maximum value $\forall h$.

\subsection{Hand gesture recognition pipeline}\label{pipeline}
\begin{figure}[t]
	\includegraphics[width=\linewidth]{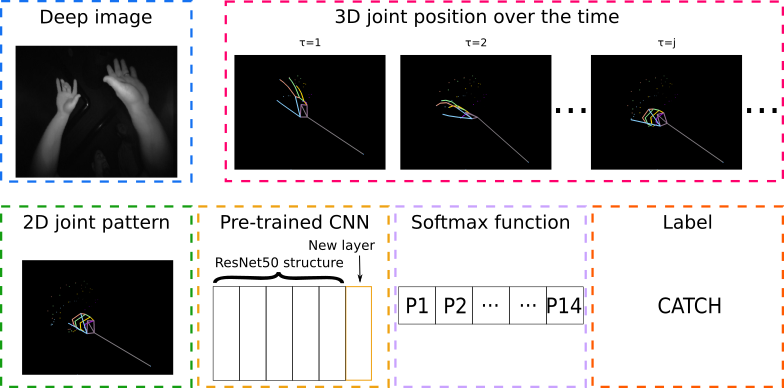}
	\caption{Pipeline of the proposed hand gesture recognition}
	\label{fig:pipeline}
\end{figure}

We propose a view-based approach able to describe the performed movement over time whose pipeline is illustrated in \figurename \ref{fig:pipeline}.
As input, a user performs different gestures recorded as depth images by using a Leap Motion sensor (blue box). A 3D gesture visualization containing temporal information is created by using the joint positions of the 3D skeleton of the hands (magenta box) as obtained from the Leap Motion sensor. From the 3D environment, we create a 2D image projecting the obtained 3D points on a view plane (green box). The created image is fed to the pre-trained convolutional neural network (yellow box), to whose output neurons (14 such as the gesture classes to be classified) a softmax function is applied which generates a probability distribution which finally represents the predicted classes (purple box). Finally, the gesture is labeled with the class that obtains the maximum probability value (orange box). In the following the two main steps are described.


\subsubsection{The 3D visualizer}
We used the VisPy\footnote{http://vispy.org} library to visualize the 3D data of the hands in a programmable 3D environment. The visualizer is able to acquire the skeleton data both from the files belonging to the LMDHG dataset (through the Pandas\footnote{https://pandas.pydata.org} library), and in real time using the Leap Motion SDK wrapped through the popular framework ROS (Robot Operating System) \cite{ros} which provides a convenient publish/subscribe environment as well as numerous other utility packages.

A 3D hand skeleton is created by exploiting the tracking data about each finger of the hand, the palm center, the wrist and the elbow positions. If at a certain time the whole or a part of a finger is not visible, the Leap Motion APIs allows to estimate the finger positions relying on the previous observations and on the anatomical model of the hand.

\begin{figure}[t]
	\subfigure[Catching\label{catching}]{	
			\includegraphics[width=0.5\linewidth]{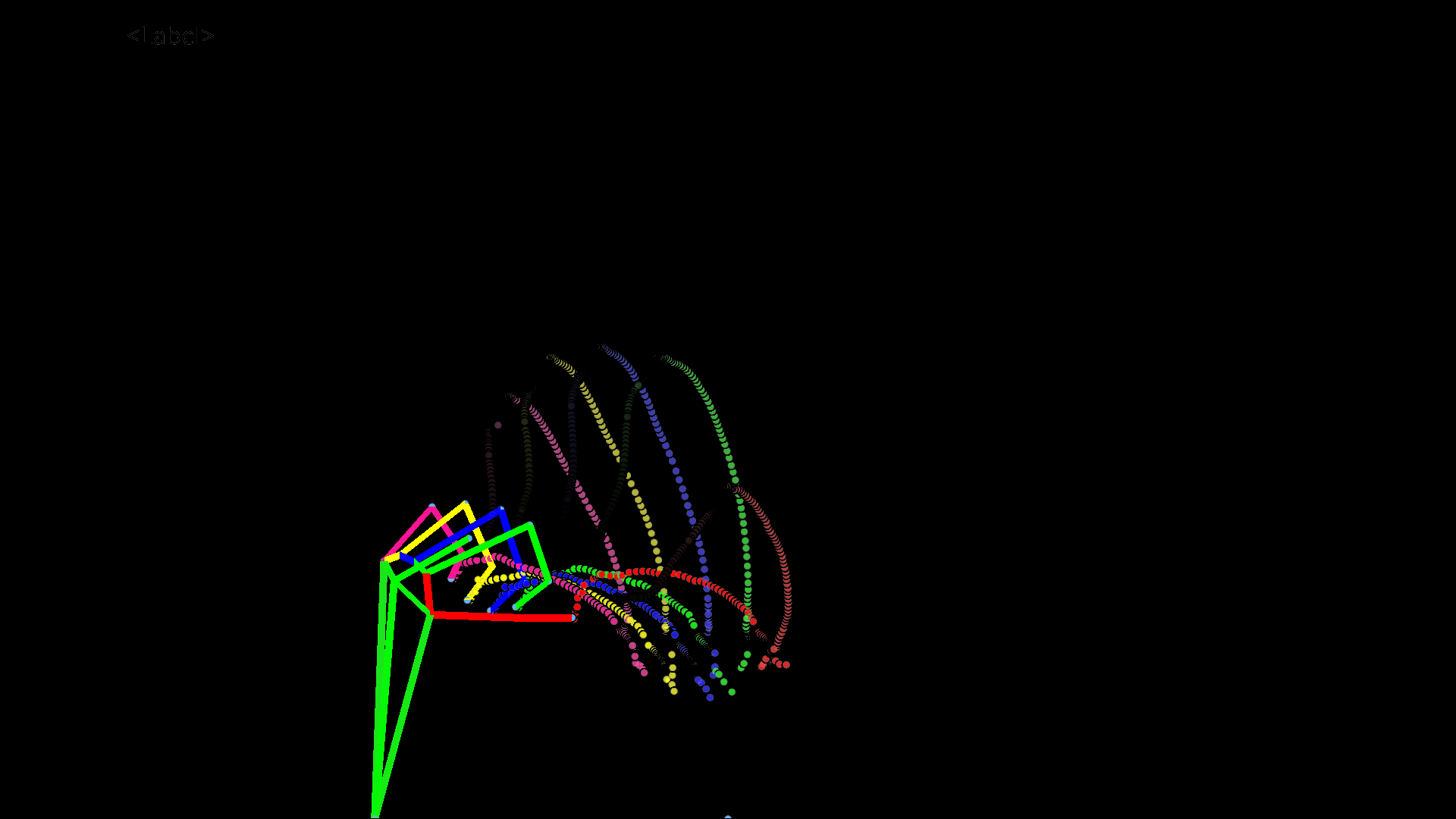}}\quad
	\subfigure[Rotating\label{rotating}]{	
			\includegraphics[width=0.5\linewidth]{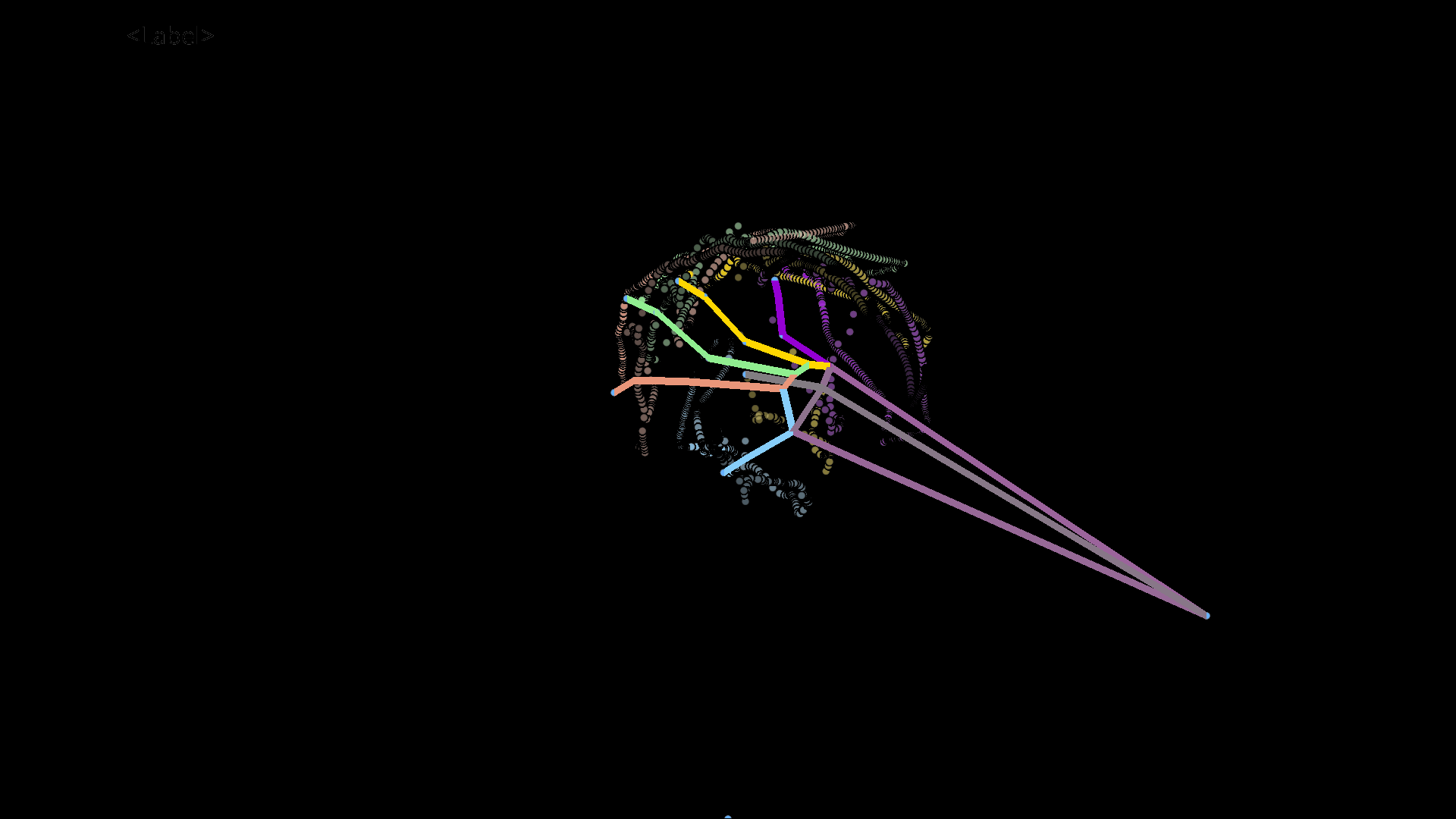}}
	\subfigure[Scroll\label{scroll}]{	
			\includegraphics[width=0.5\linewidth]{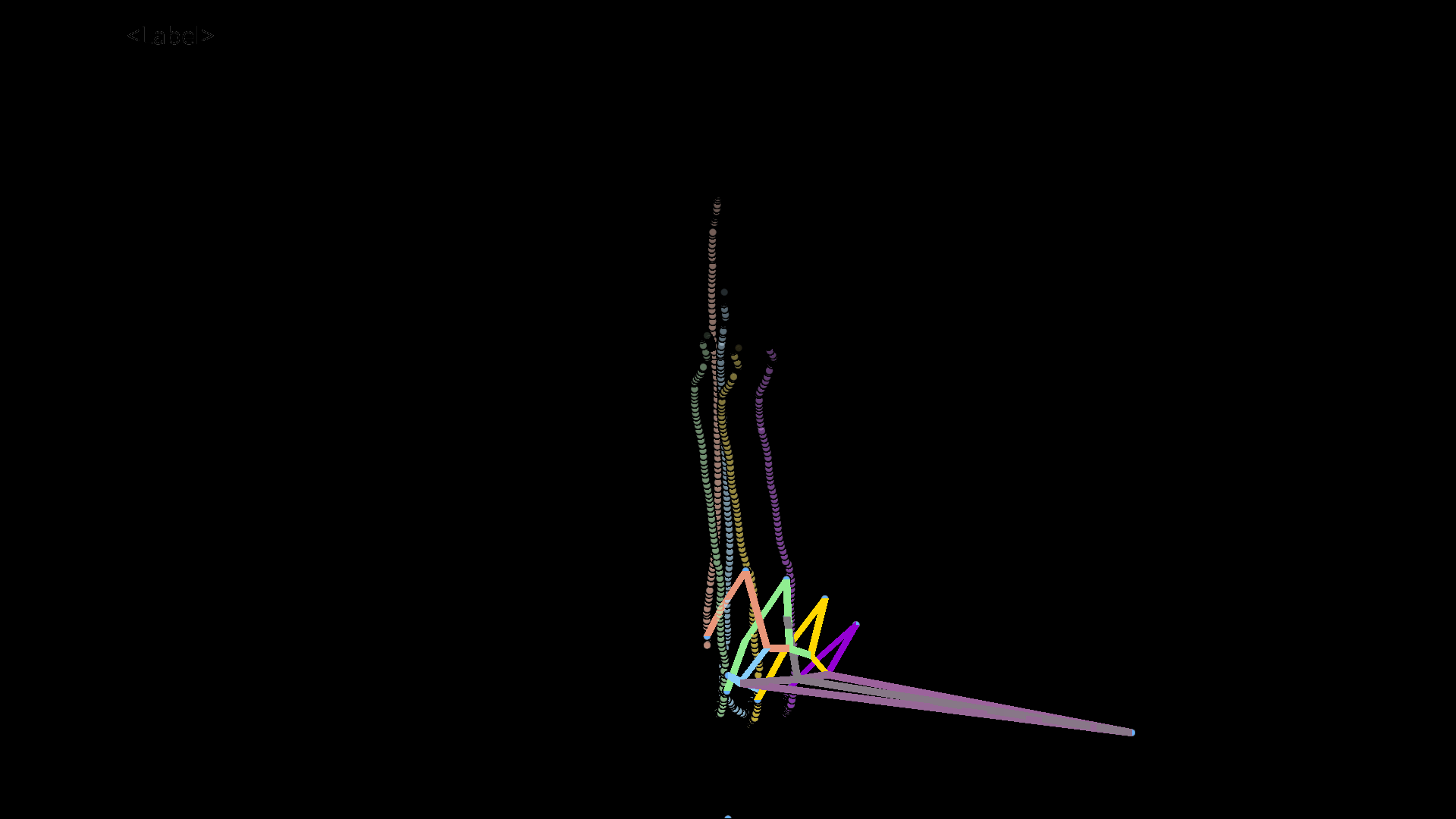}}\quad
	\subfigure[Shaking\label{shaking}]{	
			\includegraphics[width=0.5\linewidth]{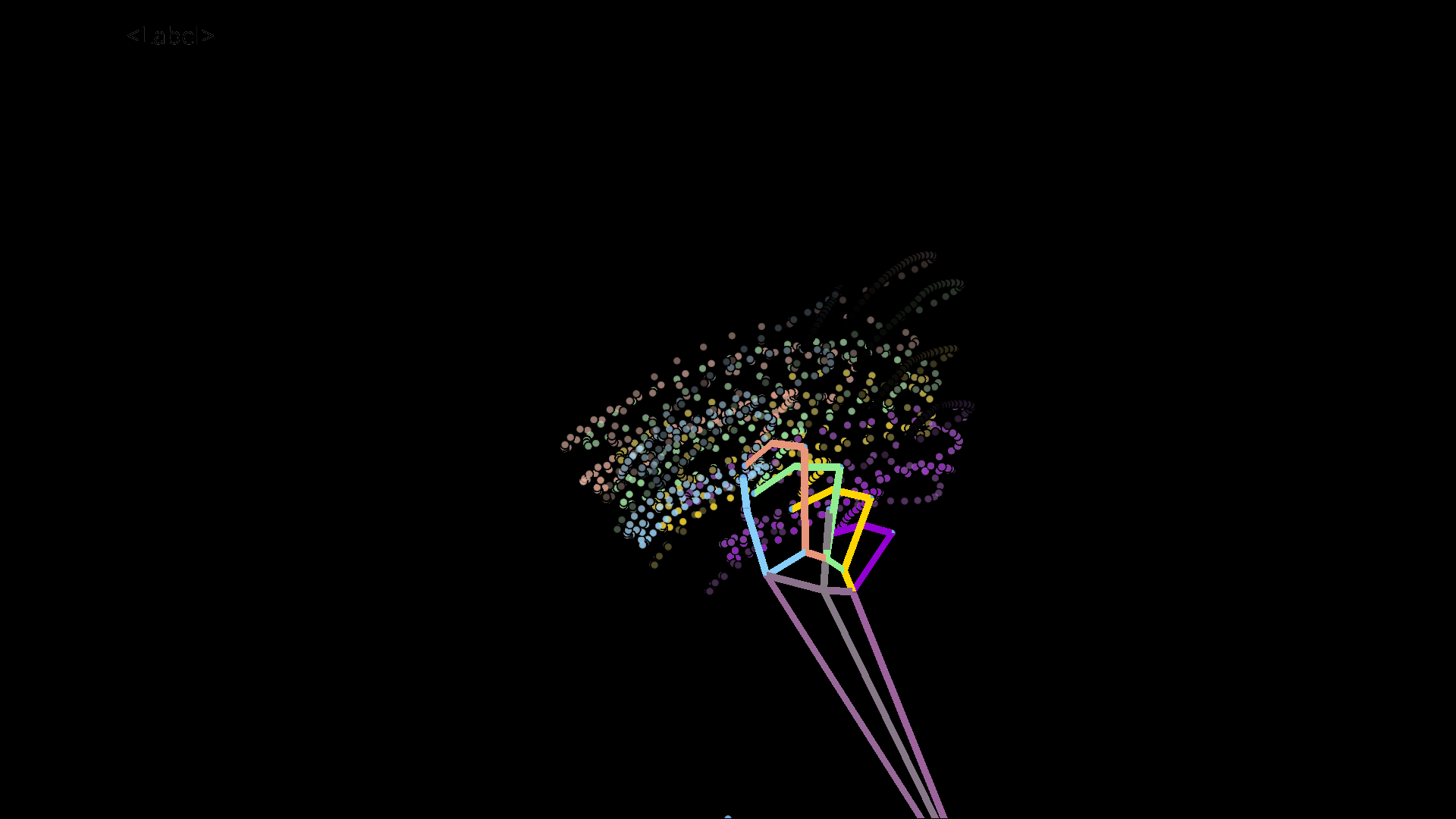}}
	\subfigure[Draw line\label{Line}]{	
			\includegraphics[width=0.5\linewidth]{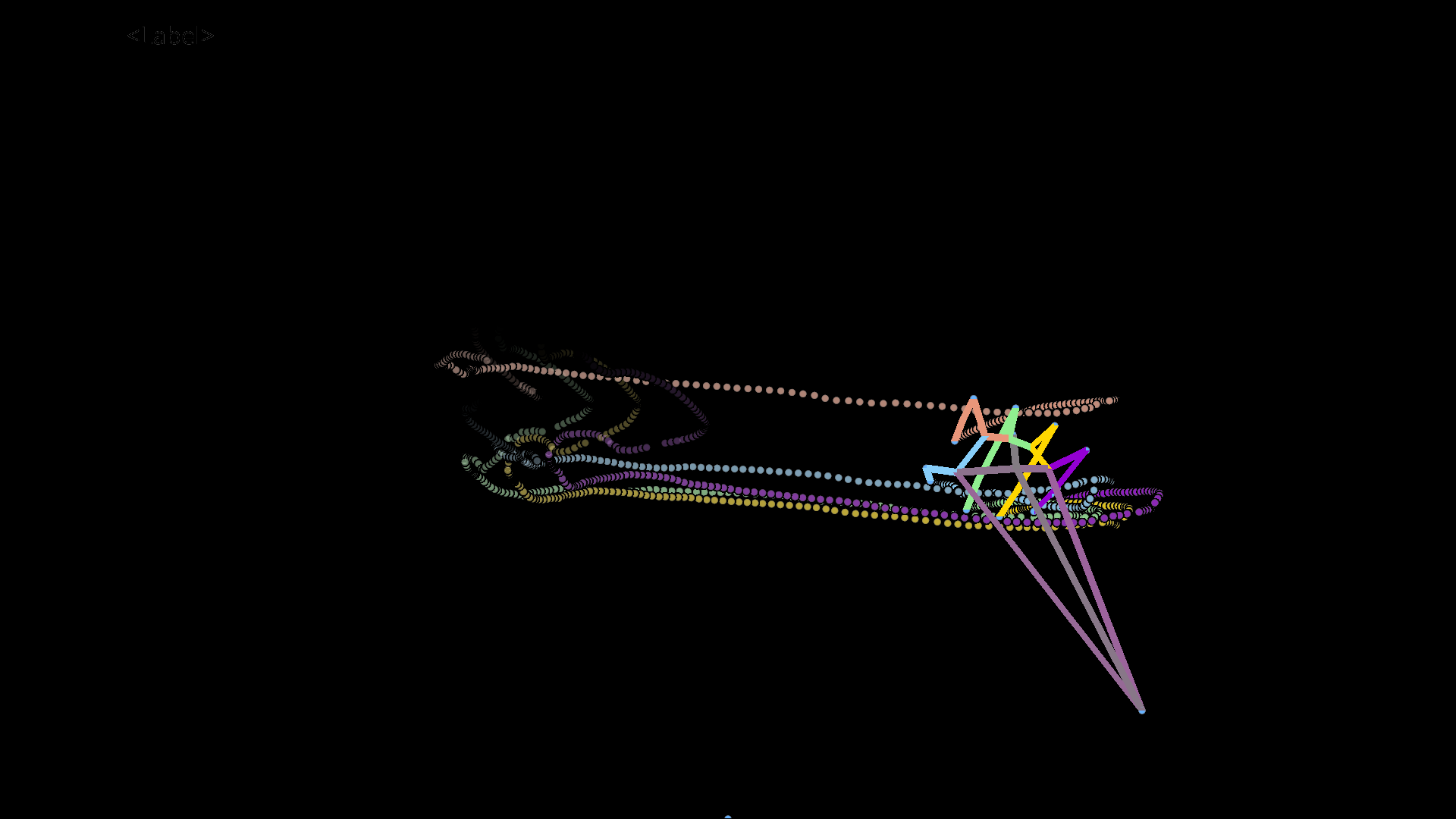}}\quad
	\subfigure[Zoom\label{Zoom}]{	
			\includegraphics[width=0.5\linewidth]{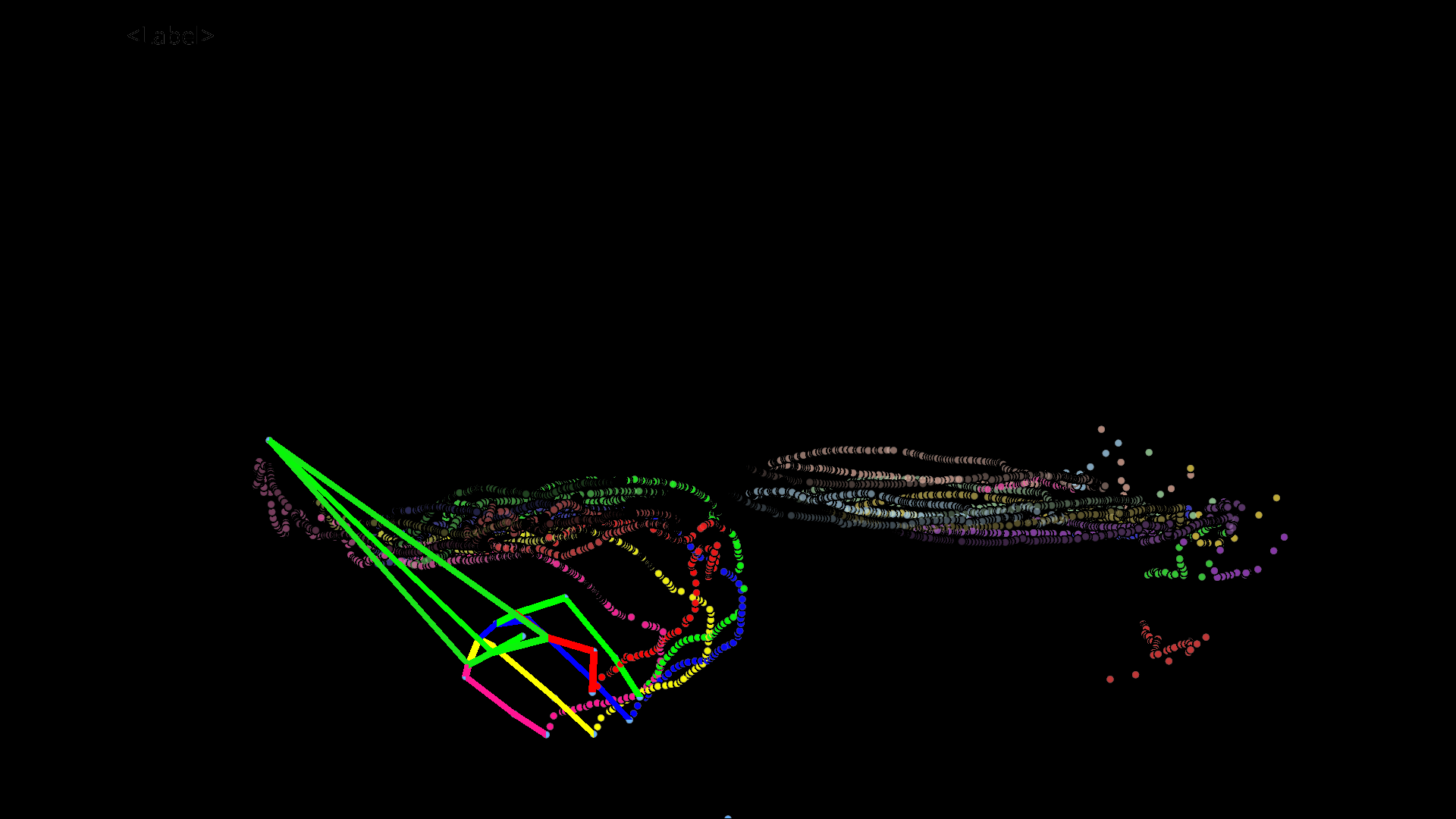}}
	\caption{Examples of 2D hand gesture patterns}
	\label{fig:gesturePattern}
\end{figure}

Once the 3D joint positions are acquired, spatial and temporal information of each gesture movement are encoded by creating a 3D joint gesture image, where 3D points and edges are depicted in the virtual space for each finger. Here, the color intensity of the joints representing the fingertips changes at different time instants; specifically, recent positions ($\tau \sim T_{i}$) have more intense colors, while earlier positions ($\tau \sim 0$) have more transparent colors. 
Finally, we create a 2D image by projecting the 3D points obtained at the last instant of the gesture on a view plane. In particular, we project the 3D fingertips of the hands on a plane corresponding to the top view, which represents hands in a ``natural" way as a human usually see them. 
\figurename \ref{fig:gesturePattern} shows an examples of the 2D hand gesture patterns obtained for four different gestures.
Although this view does not contain all the information available in the 3D representation of the hands, we have found that it is sufficient for a CNN to classify the set of dynamic gestures under study very accurately.


\subsubsection{Classification method}
The proposed method leverages a pre-trained ResNet-50 \cite{resnet}, a state-of-the-art 2D CNN that has been modified and fine-tuned to classify the images produced by our 3D visualizer. We decided to use a ResNet-50 because this kind of architecture is pre-trained on ImageNet \cite{imagenet} and it is one of the fastest at making inference on new images, having one of the lowest FLOPS count among all the architectures available today \cite{cnn_benchmarking}. Unfortunately, given the modest size of the original LMDHG dataset, it would not have been possible to train from scratch a 3D CNN model capable of classifying all the available information coming from the LM sensor.


\subsection{The LMDHG gestures dataset}\label{DatasetDescription}
Most of the reviewed gesture datasets are composed of gestures executed with a single hand, performed and recorded perfectly, with no noise or missing parts, and segmented always with the same duration. These hypotheses ensure a good class separation improving the classification results but they are far from the reality. For instance, it is not unusual to record hand trembles during the gestures including a significant amount of noise.
 
To improve the evaluation of different methods over a more realistic dataset, Boulahia et al. \cite{boulahia2017dynamic} define a dataset of unsegmented sequences of hand gestures performed both with one and two hands. At the end of each gesture, the involved participants were asked to perform a ``rest" gesture, i.e. keeping the hands in the last movement position for a few seconds, thus providing a kind of \textit{null gesture} that can be used to recognize the ending of a certain movement.

We chose their dataset as a starting point to test our method because it was the most realistic dataset created using the Leap Motion sensor that we were able to identify. It is our opinion that the original LMDHG paper provides three major contributions: i) the evaluation of the method proposed by the authors against the DHG dataset \cite{de2016skeleton}, ii) the evaluation of the method proposed by the authors against the properly segmented version of the LMDHG dataset, iii) the evaluation of the method proposed by the authors against the non-segmented version (i.e. without providing their classifier with the truth value on where a gesture ends and the next one starts) of the LMDHG dataset. For this paper, we decided to apply our method in order to replicate and improve only point ii), namely, against the properly segmented LMDHG dataset.

This dataset contains 608 ``active" plus 526 ``inactive" (i.e. classified as the \textit{Rest} gesture) gesture samples, corresponding to a total of 1134 gestures. These gesture instances fall into 14 classes, \textit{Point to}, \textit{Catch}, \textit{Shake down}, \textit{Shake}, \textit{Scroll}, \textit{Draw Line}, \textit{Slice}, \textit{Rotate}, \textit{Draw C}, \textit{Shake with two hands}, \textit{Catch with two hands}, \textit{Point to with two hands}, \textit{Zoom} and \textit{Rest}, of which the last 5 gestures are performed with two hands.
Unfortunately, the gesture classes are divided unevenly having a number of samples not uniform among them. Indeed, most of the classes have roughly 50 samples, except \textit{Point to with hand raised} that presents only 24 samples and \textit{Rest}, as previously said, that presents 526 samples. 


\section{Experiments}\label{Experiments}
In this section, we present the experimental results obtained by processing the LMDHG dataset, represented in form of images from our 3D visualizer. The main obtained results concern the training of three distinct models through (i) images depicting a single view of the hands from above (see sub-section \ref{EvaluationSingleView}); (ii) images obtained by stitching two views together (from the top and from the right) to provide further information to the classifier (see sub-section \ref{EvaluationDoubleView}); and (iii) a new dataset that we publicly release at this URL\footnote{\url{https://imaticloud.ge.imati.cnr.it/index.php/s/YNRymAvZkndzpU1}} containing about 2000 new gestures performed more homogeneously with each other, with less noise and with fewer mislabeling occurrences than in the LMDHG dataset (see sub-section \ref{EvaluationNewDataset}). Indeed, we deem this dataset is richer and more suitable for the initial stages of training of CNN models when there are few samples available and it is important that the signal-to-noise ratio of the information used for training is high.



\subsection{Training of the models}\label{TrainingDescription}

The training took place using Jupyter Notebook and the popular deep learning library, Fast.ai \cite{fastai}, based on PyTorch. The hardware used was a GPU node of the new high-performance EOS cluster located within the University of Pavia. This node has a dual Intel Xeon Gold 6130 processor (16 cores, 32 threads each) with 128 GB RAM and 2 Nvidia V100 GPUs with 32 GB RAM.

The training was performed on 1920x1080 resolution images rendered by our 3D visualizer, properly classified in directories according to the original LMDHG dataset and divided into training and validation sets, again following the indications of the original paper \cite{boulahia2017dynamic}.

As previously mentioned, the model chosen for training is a pre-trained version of a ResNet-50 architecture. Fast.ai convenient APIs, allow to download pre-trained architecture and weights in a very simple and automatic way. Fast.ai also automatically modifies the architecture so that the number of neurons in the output layer corresponds to the number of classes of the current problem, initializing the new layer with random weights.

The training has performed using the progressive resizing technique, i.e. performing several rounds of training using the images of the dataset at increasing resolutions to speed up the early training phases, have immediate feedback on the potential of the approach, and to make the model resistant to images at different resolutions (i.e. the model generalizes better on the problem). The specific section in \cite{cellular_super_resolution} explains very well the concept of progressive resizing. For our particular problem, we have chosen the resolutions of 192, 384, 576, 960, 1536 and 1920 px (i.e. 1, 2, 3, 5, 8 and 10/10 of the original 1920x1080 px resolution).

Each training round at a given image resolution is divided into two phases (\textit{a} = frozen, \textit{b} = unfrozen), each consisting of 10 training epochs. In phase \textit{a}, the weights of all the layers of the neural network except those of the new output layer are frozen and therefore are not trained (they are used only in the forward pass). In phase \textit{b}, performed with a lower learning rate (LR), typically of one or two orders of magnitude less\footnote{Fast.ai's \textit{Learner} class has a convenient \textit{lr\_find()} method that allows to find the best learning rate with which to train a model in its current state.}, all layers, even the convolutional ones, are trained to improve the network globally.

As neural network model optimizer, we chose Ranger as it combines two of the best state-of-the-art optimizers, RAdam \cite{RAdam} (Rectified Adam) and Lookahead \cite{Lookahead}, in a single optimizer. Ranger corrects some inefficiencies of Adam \cite{Adam}, such as the need for an initial warm-up phase, and adds new features regarding the exploration of the loss landscape, keeping two sets of weights, one updated faster and one updated more slowly, and interpolating between them to improve the convergence speed of the gradient descent algorithm.

Once all the training rounds were completed, the model with the best accuracy was selected for the validation phase. At the same accuracy between checkpoints at different training rounds, the model generated by the lowest round (i.e. trained with lower image resolution) was selected and reloaded for validation. This has a substantial advantage in the inference phase since smaller images are classified faster.

All the code and jupyter notebooks described in this section are available at the following URL\footnote{https://github.com/aviogit/dynamic-hand-gesture-classification}.

\subsection{Evaluation on the LMDHG gesture dataset - single view}\label{EvaluationSingleView}
To allow a further comparison with the method provided by Boulahia et al. \cite{boulahia2017dynamic}, we split the dataset according to their experiments, i.e. by using sequences from 1 to 35 of the dataset to train the model (779 samples representing $\sim70\%$ of the dataset) and sequences from 35 to 50 to test it (355 samples representing $\sim30\%$ of the dataset).

\begin{figure}[htbp]
	\centering
	\includegraphics[height=0.45\textheight]{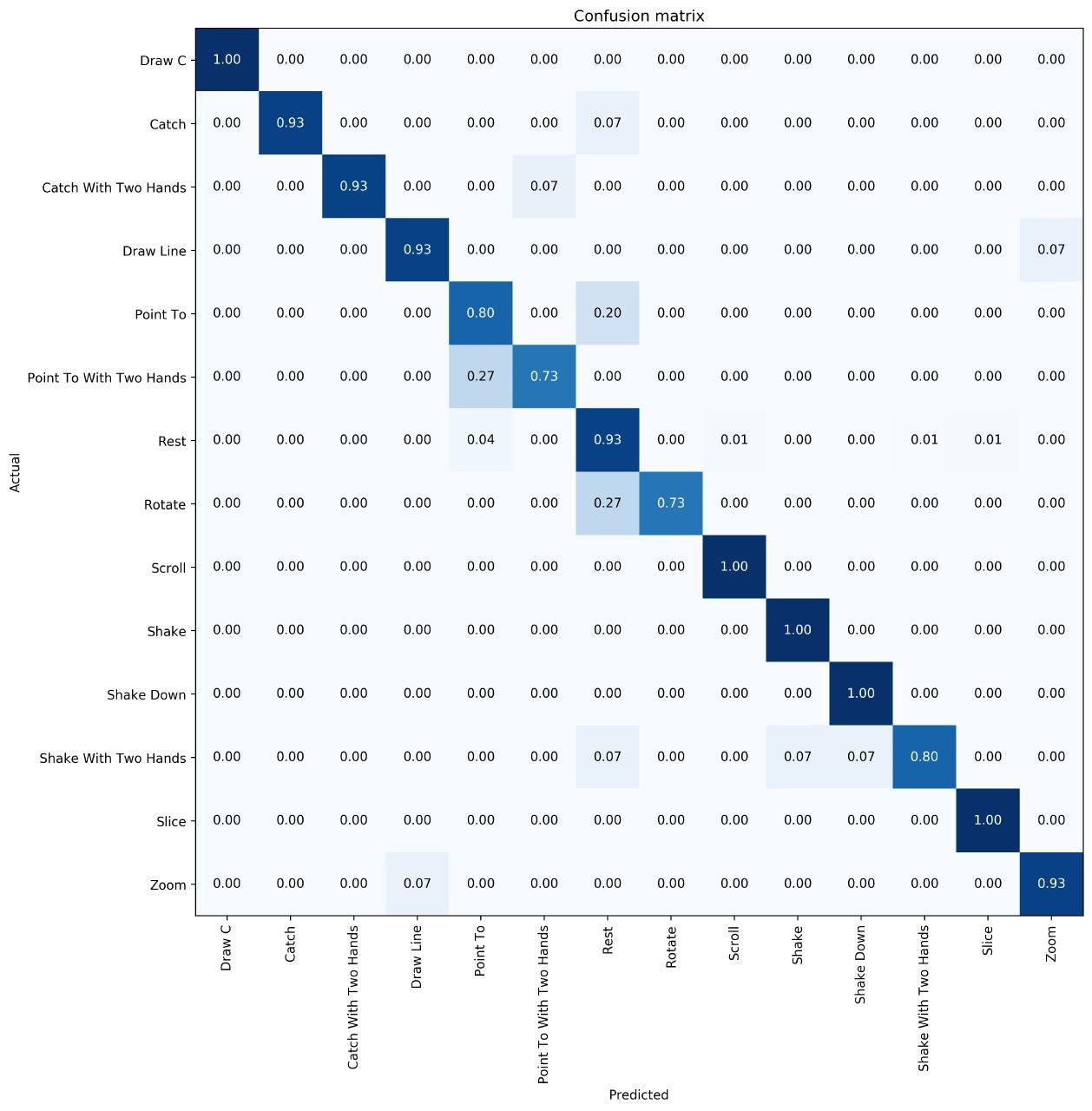}
	\caption{Confusion matrix obtained using a single view.}
	\label{fig:ConfMatrixOneView}
\end{figure}

With this partition, our approach reaches an accuracy of 91.83\% outperforming the 84.78\% performed by Boulahia et al.
From the confusion matrix illustrated in \figurename\ \ref{fig:ConfMatrixOneView}, we can notice that most of the classes are well recognized with an accuracy over 93\%. 
Misclassifications occur when the paired actions are quite similar. For example,
the gestures \textit{Point to} and \textit{Rotate}, which are recognized with an accuracy of 80\% and 73\% respectively, are confused with the nosiest class \textit{Rest}; \textit{Point to with two hands}, recognized with an accuracy of 73\%, is confused with the close class \textit{Point to}; while \textit{Shake with two hands}, recognized with an accuracy of 80\%, is reasonably confused with the two close classes \textit{Shake}, \textit{Shake down}.

\begin{figure}[htbp]
	\centering
	\includegraphics[width=\linewidth]{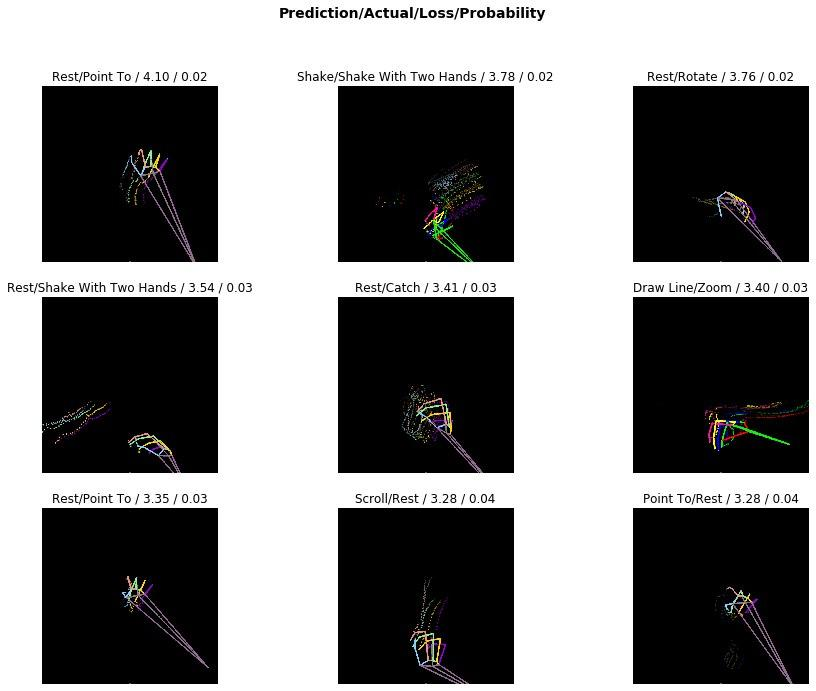}
	\caption{Top losses plot obtained using a single view.}
	\label{fig:TopLossesOneView}
\end{figure}

For a comprehensive evaluation, in \figurename\ \ref{fig:TopLossesOneView} we show the top losses for our model. The top losses plot shows the incorrectly classified images on which our classifier errs with the highest loss. In addition to the most misclassified classes deduced also from the confusion matrix analysis (i.e. \textit{Point to}, \textit{Rotate} and \textit{Shake with two hands}), from the analysis of the top losses plot, we can pinpoint a few mislabeled samples.
For example, in \figurename\ \ref{fig:TopLossesOneView} it can be seen that the third sample (prediction: \textit{Rest}, label: \textit{Rotate}) does not actually represent a \textit{Rotate} at all. The same is valid for \textit{Draw Line/Zoom}, \textit{Scroll/Rest} and \textit{Point to/Rest} samples, and having so few samples in the dataset, these incorrectly labeled samples lower the final accuracy of the model and prevent it from converging towards the global optimum.


\newpage
\subsection{Evaluation on the LMDHG gesture dataset - double view}\label{EvaluationDoubleView}

\begin{figure}[htbp]
	\centering
	\subfigure[Catching\label{DoubleCatching}]{	
		\includegraphics[width=0.35\linewidth,angle =90]{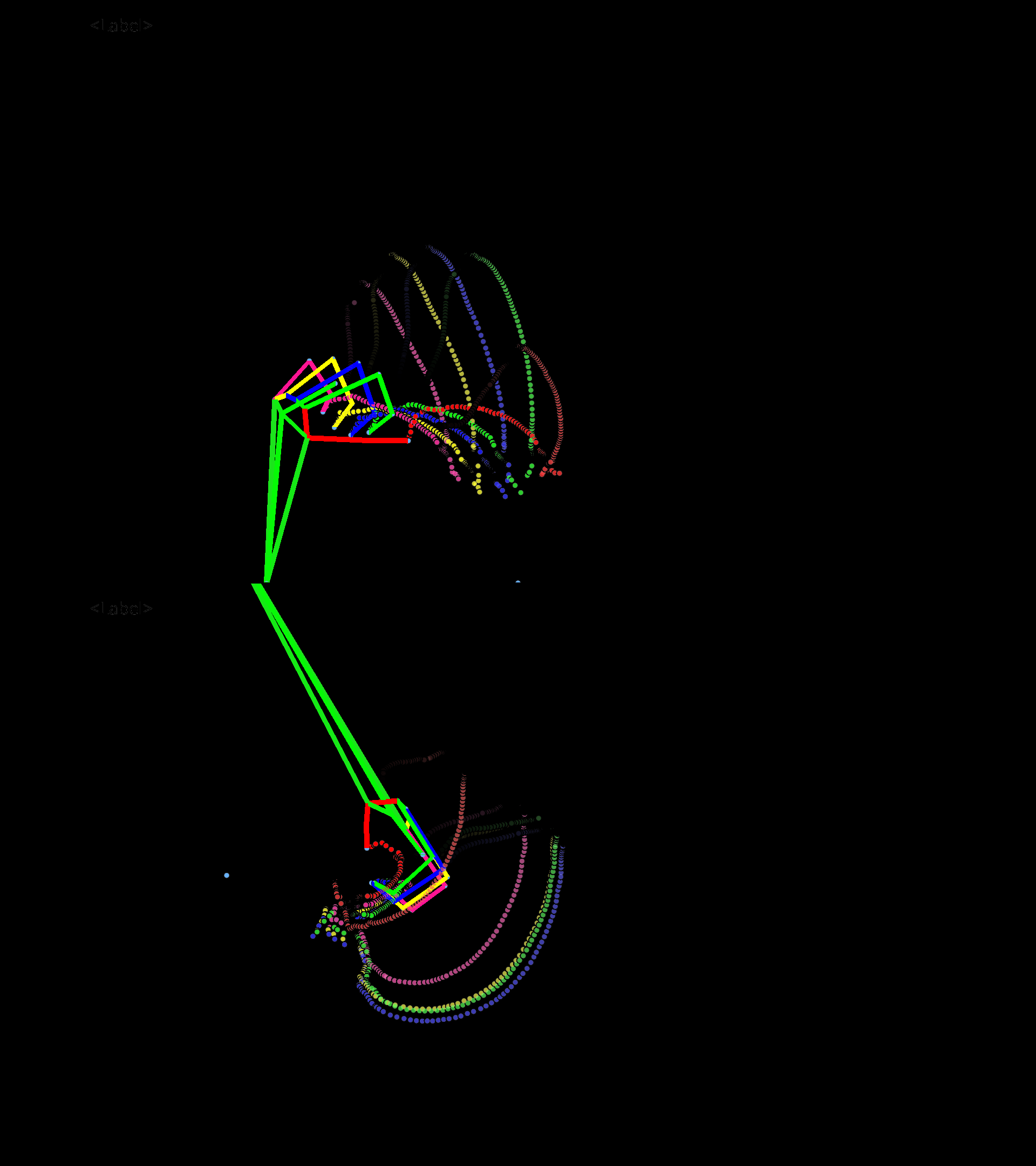}}\quad\quad
	\subfigure[Rotating\label{DoubleRotating}]{	
		\includegraphics[width=0.35\linewidth,angle =90]{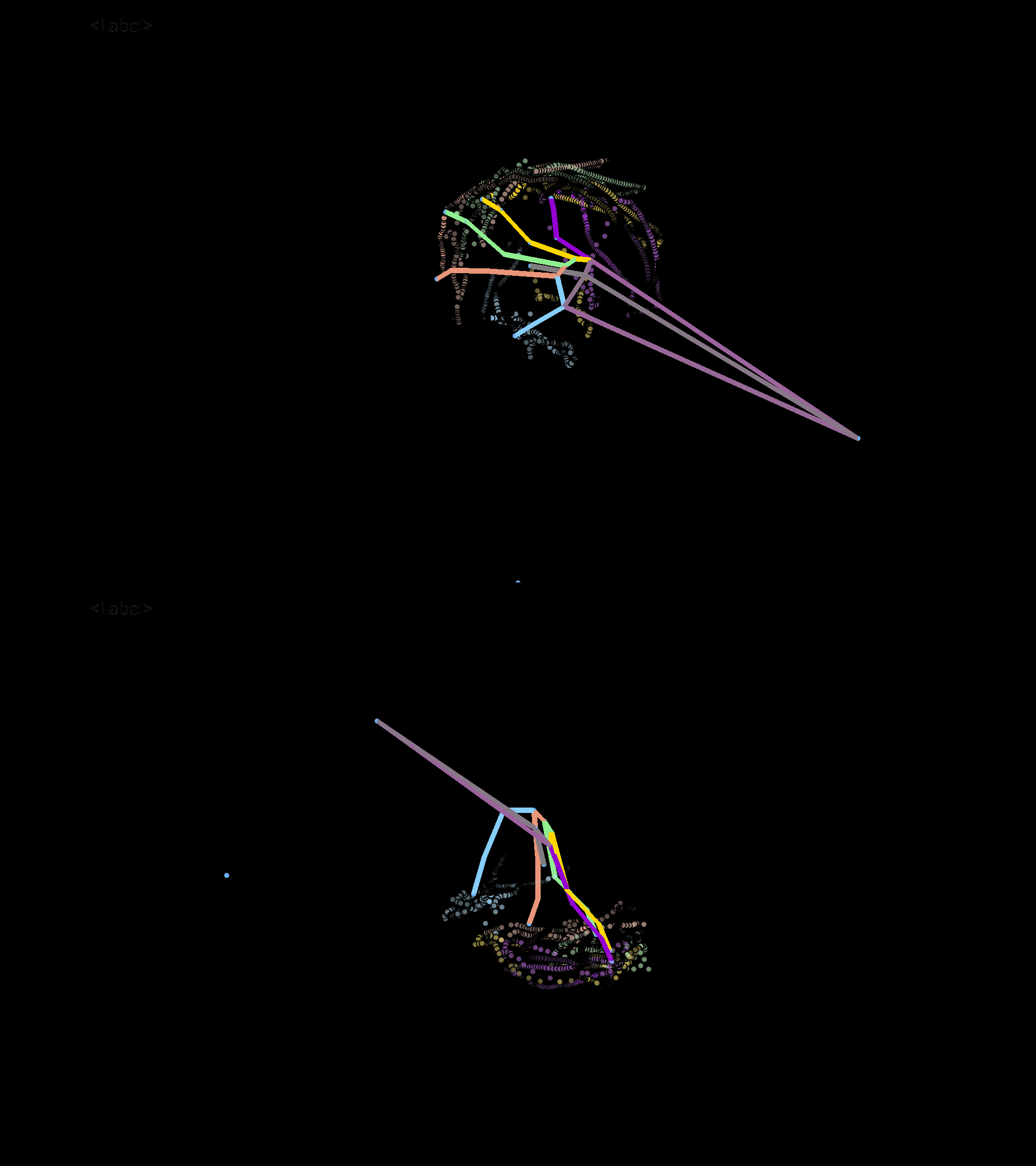}}
	\subfigure[Scroll\label{DoubleScroll}]{	
		\includegraphics[width=0.35\linewidth,angle =90]{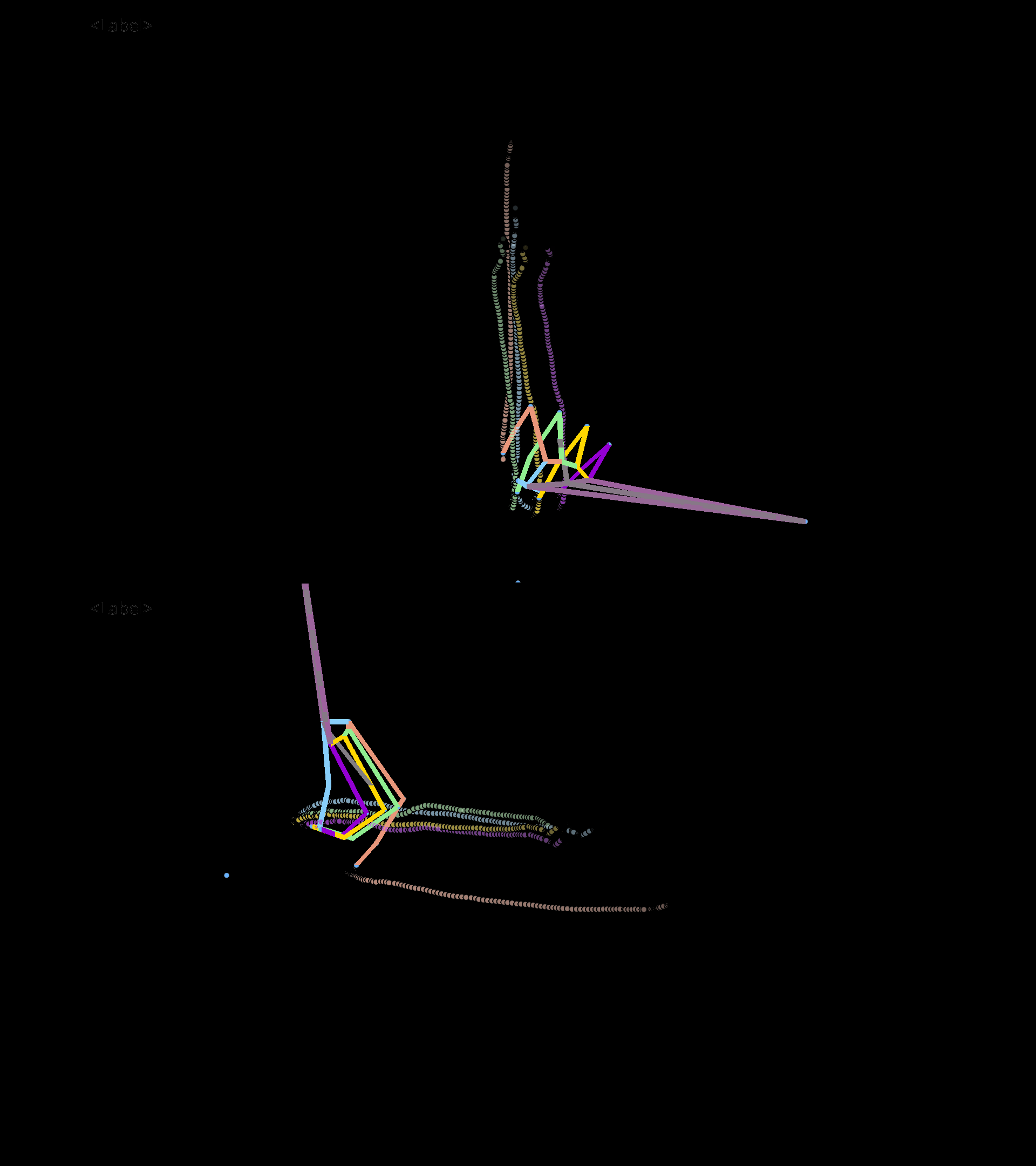}}\quad\quad
	\subfigure[Shaking\label{DoubleShaking}]{	
		\includegraphics[width=0.35\linewidth,angle =90]{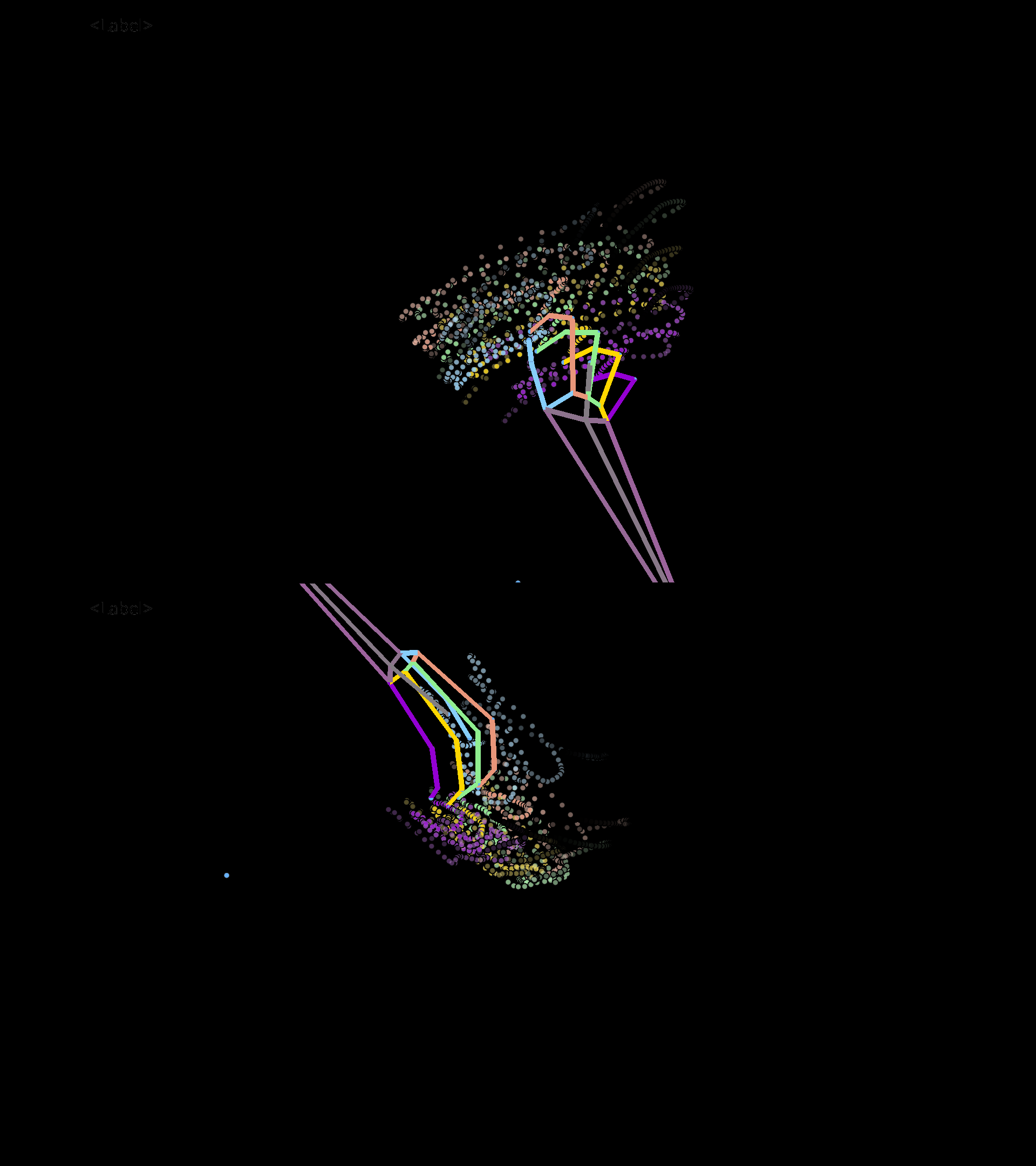}}
	\subfigure[Draw line\label{DoubleLine}]{	
		\includegraphics[width=0.35\linewidth,angle =90]{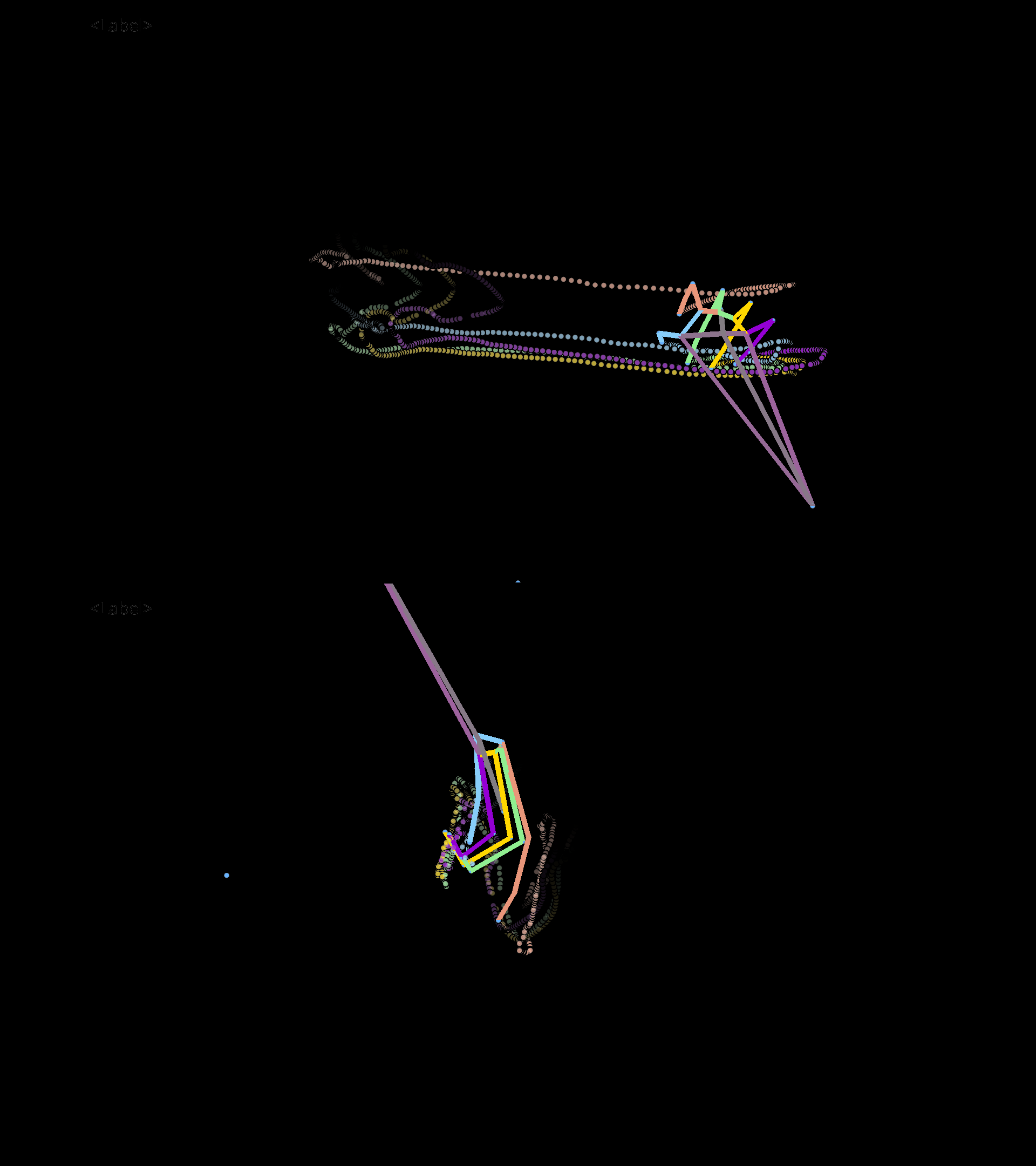}}\quad\quad
	\subfigure[Zoom\label{DoubleZoom}]{	
		\includegraphics[width=0.35\linewidth,angle =90]{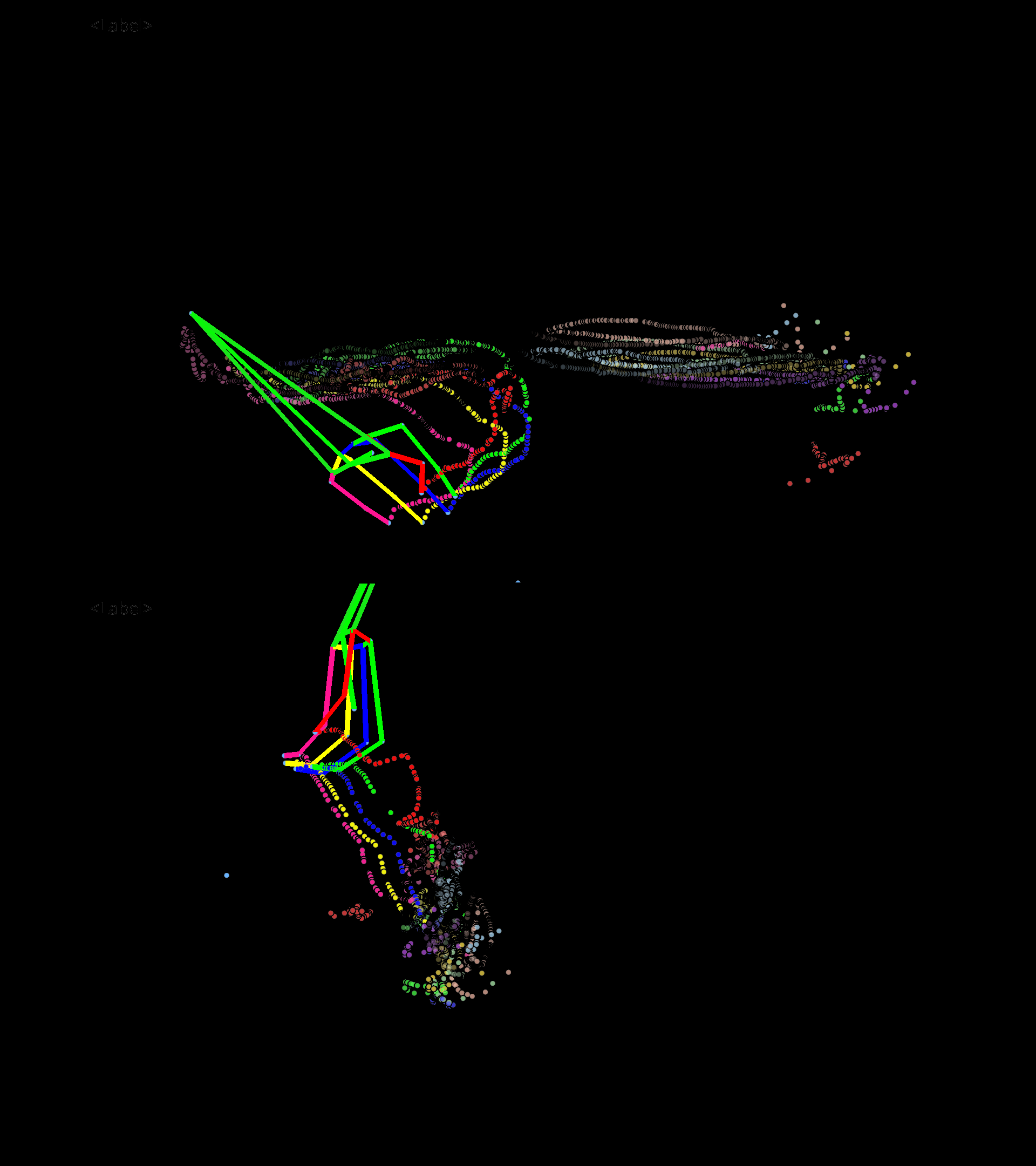}}
	\caption{Examples of 2D hand gesture patterns obtained using a double view.}
	\label{fig:gesturePatternDoubleView}
\end{figure}

To reduce these misclassifications, we trained a new model by increasing the amount of information available in each individual image: in this case, in addition to the top view, we stitch the final image by adding a view from the right. This approach allows the classifier (exactly like a human being) to disambiguate between gestures that have a strong informative content on the spatial dimension implicit in the top view (such as the \textit{Scroll} gesture, for example). Some example images are shown in \figurename\ \ref{fig:gesturePatternDoubleView}.

\begin{figure}[htbp]
	\subfigure[\label{Point1}]{	
		\includegraphics[width=0.3\linewidth]{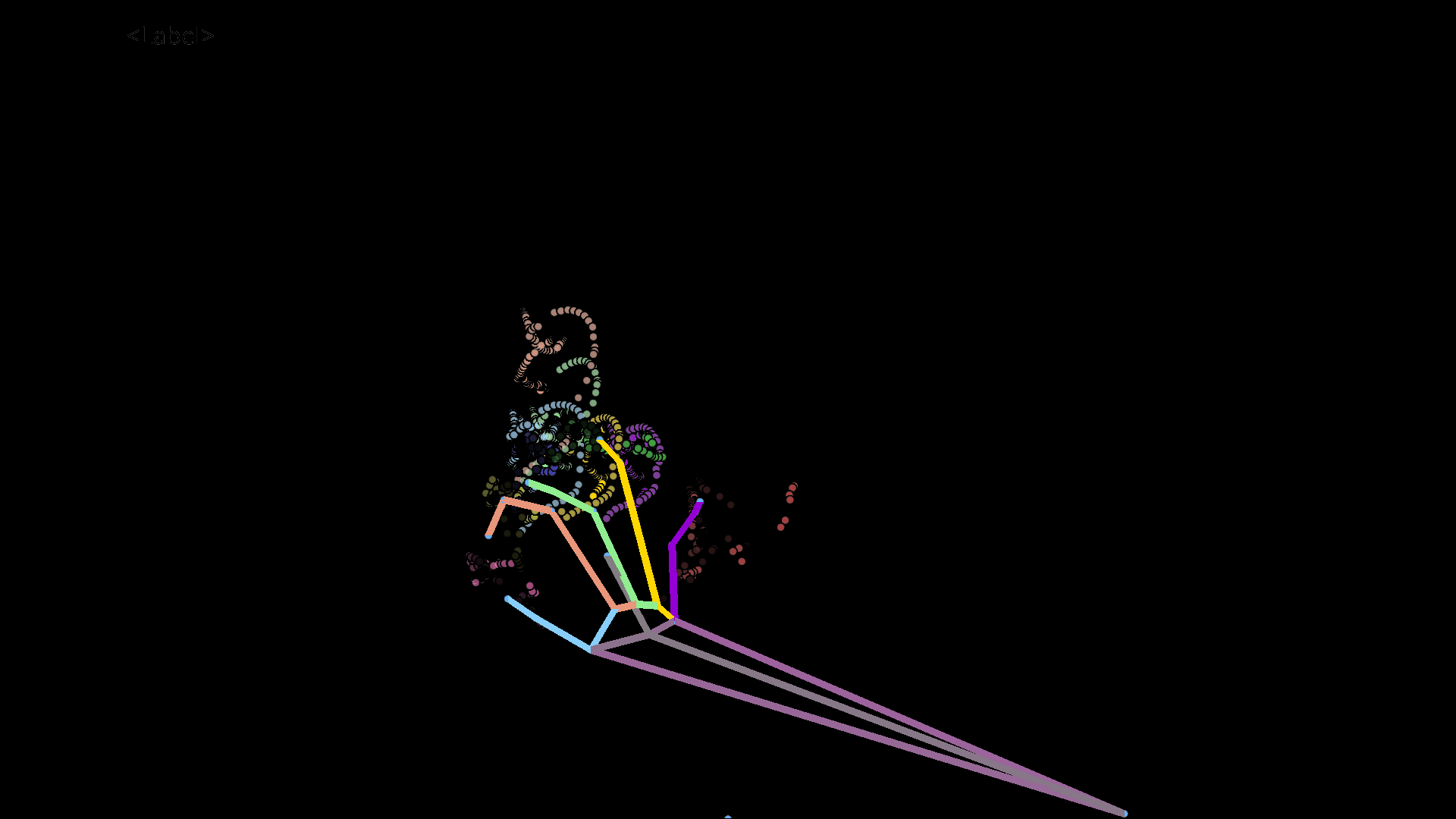}}\quad
	\subfigure[\label{Point2}]{	
		\includegraphics[width=0.3\linewidth]{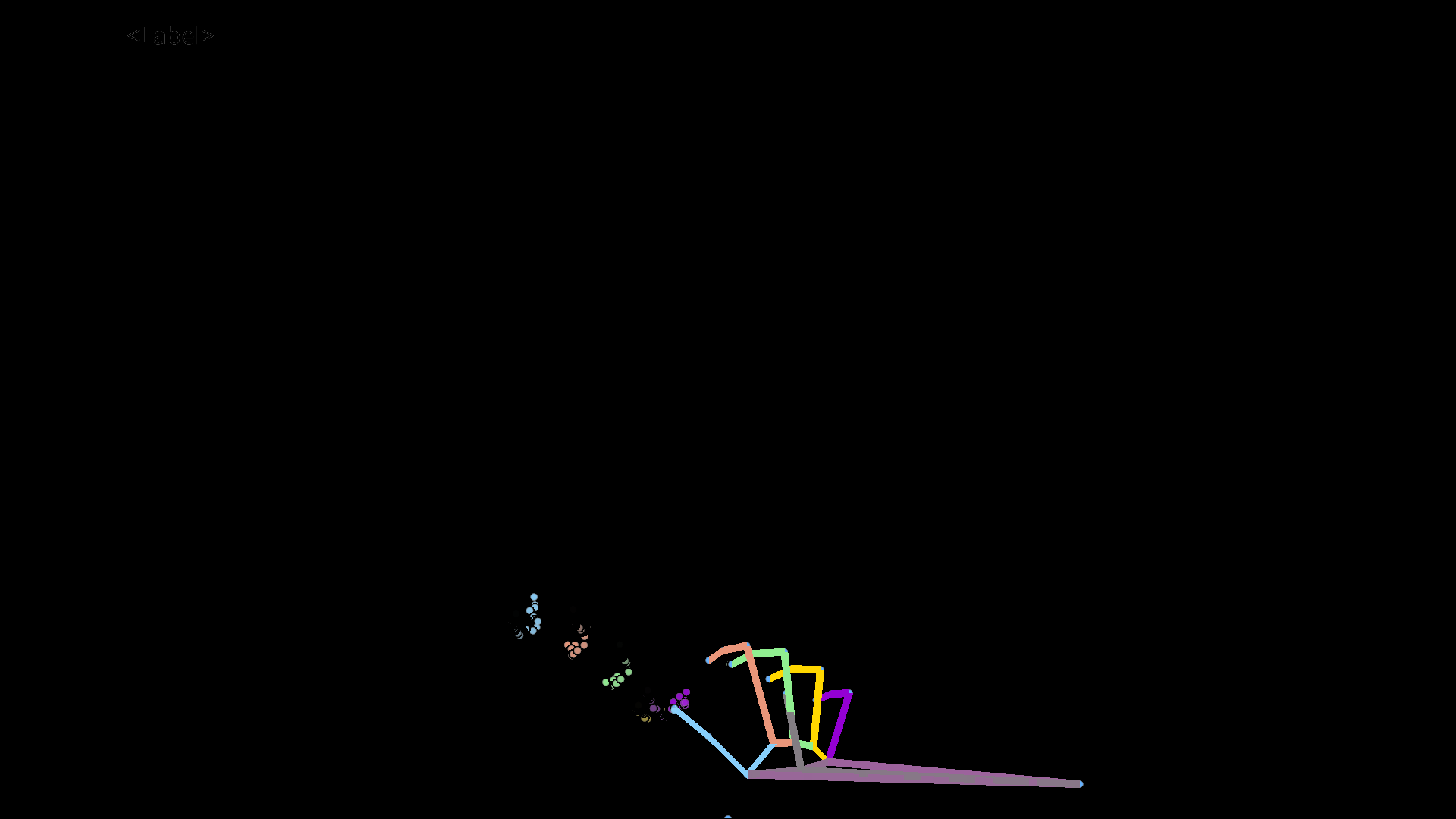}}\quad
	\subfigure[\label{Point3}]{	
		\includegraphics[width=0.3\linewidth]{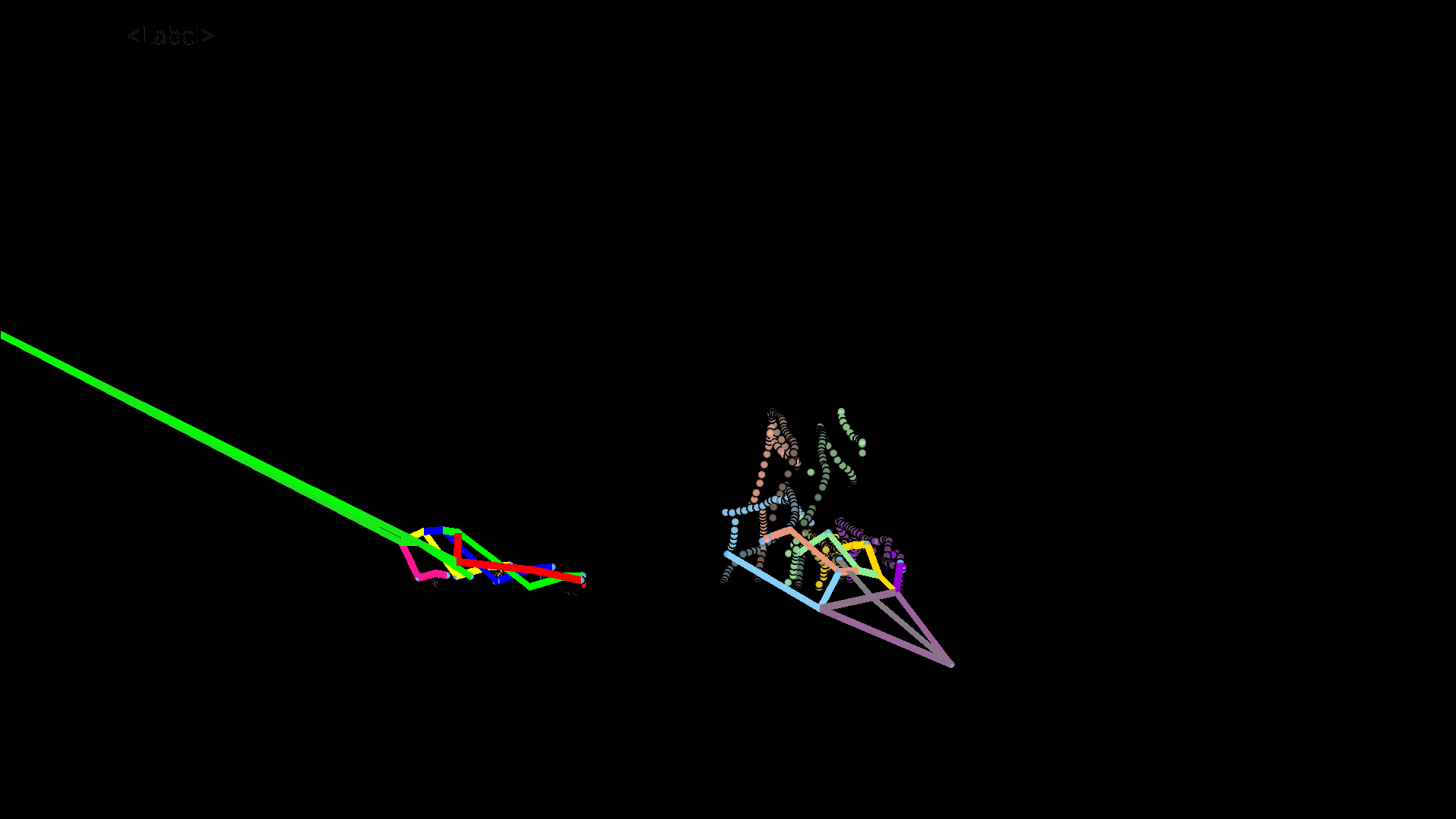}}
	\subfigure[\label{Point4}]{	
		\includegraphics[width=0.3\linewidth]{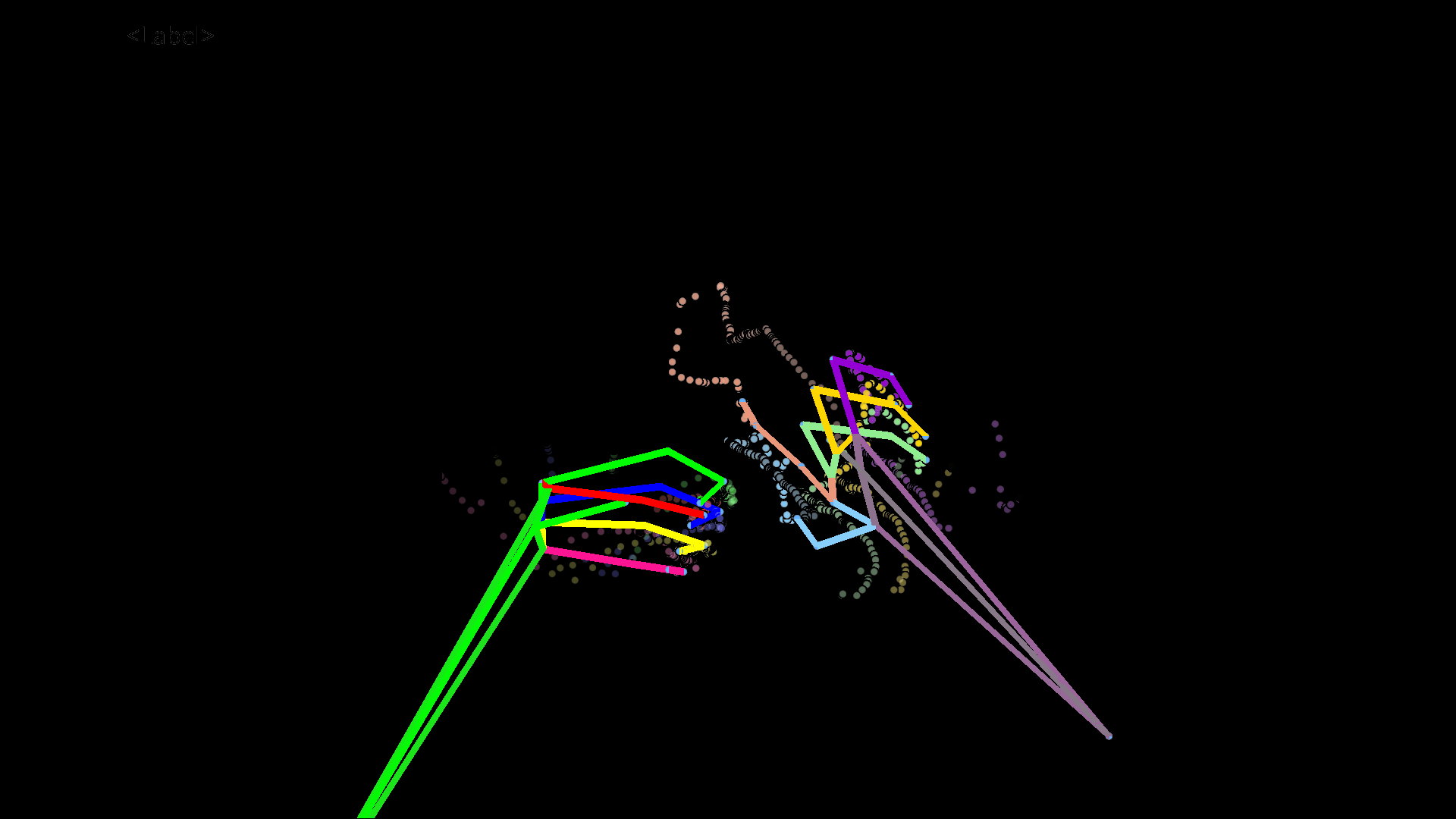}}\quad
	\subfigure[\label{Point5}]{	
		\includegraphics[width=0.3\linewidth]{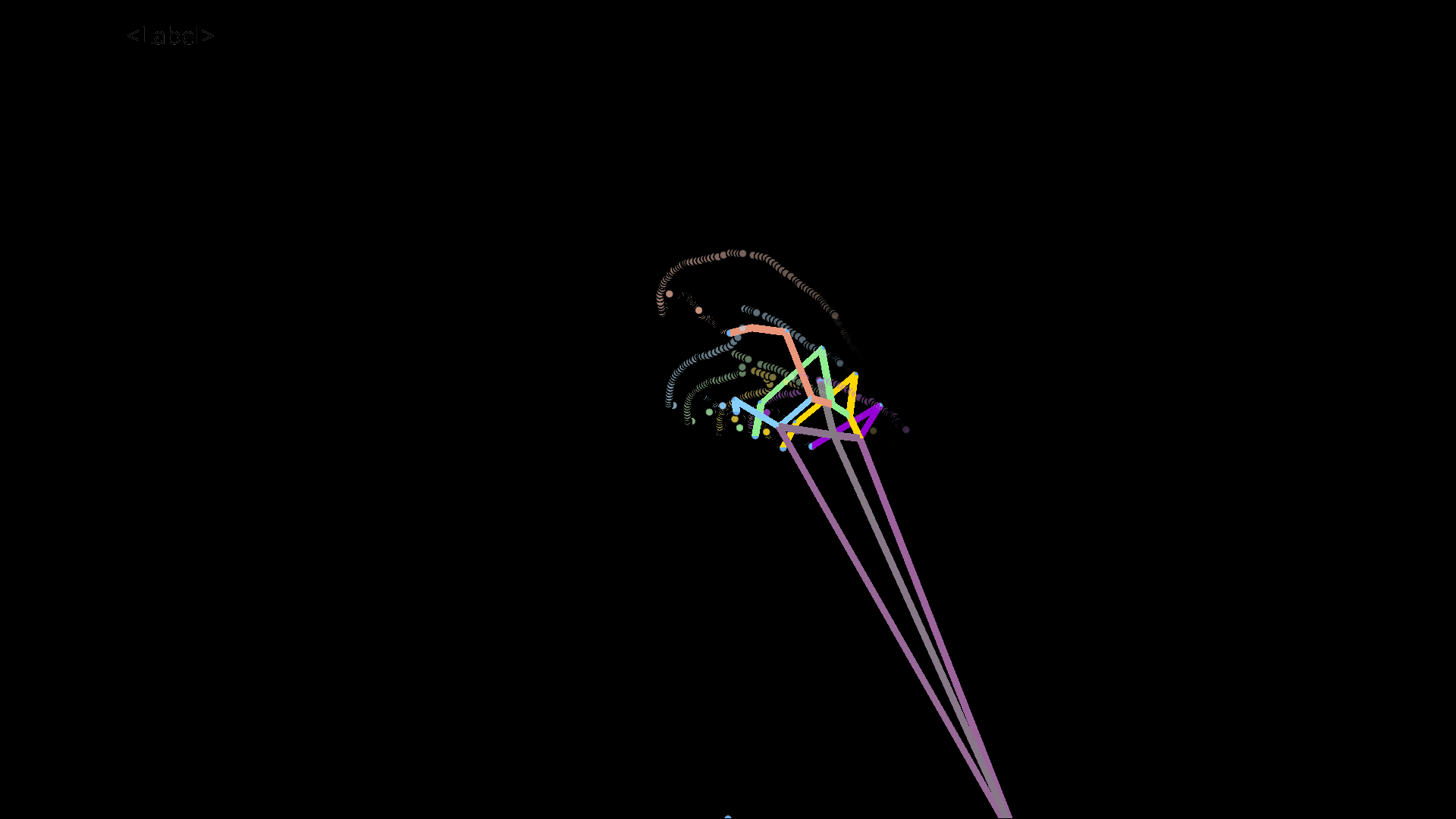}}\quad
	\subfigure[\label{Point6}]{	
		\includegraphics[width=0.3\linewidth]{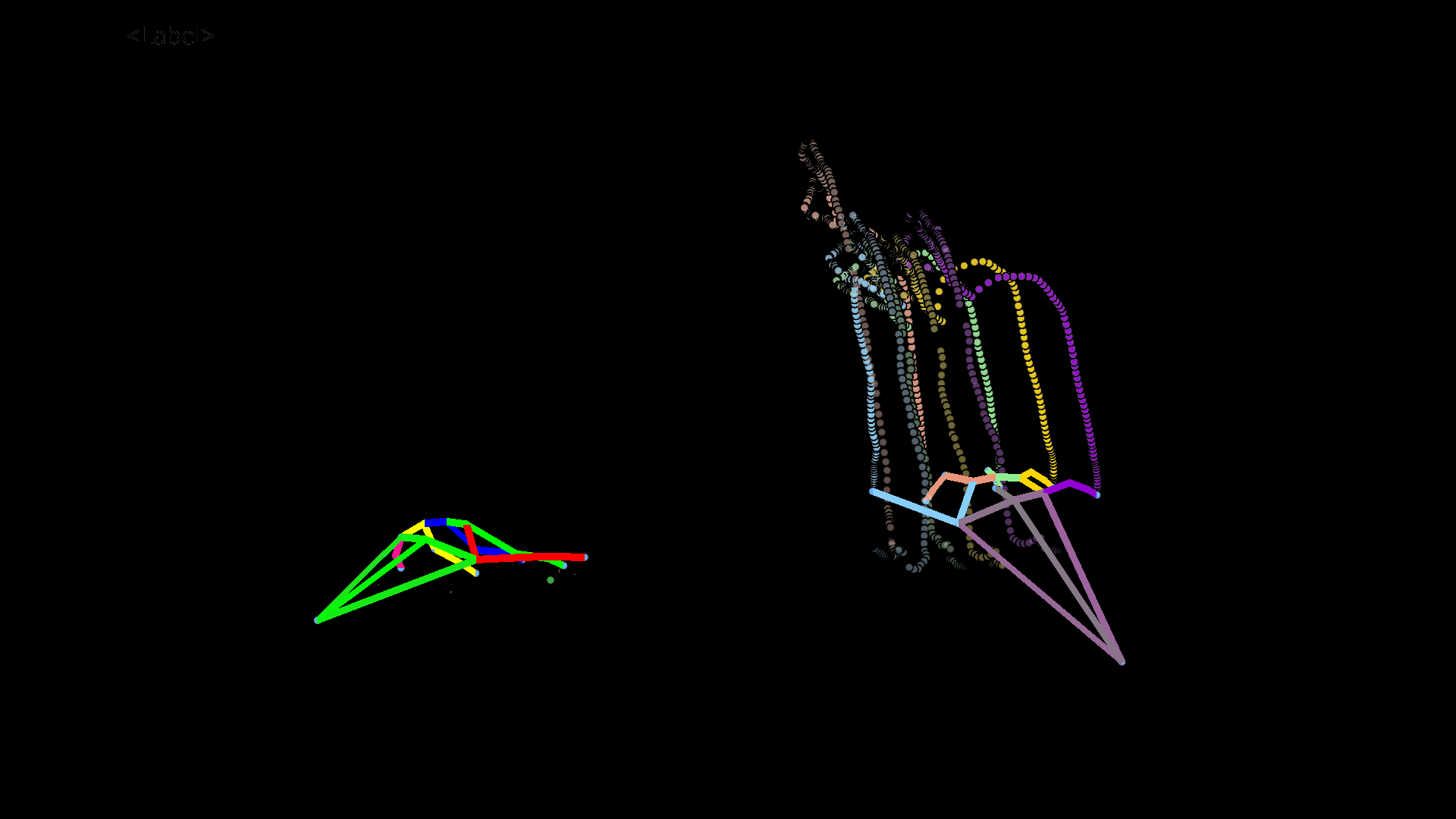}}
	\subfigure[\label{Point7}]{	
		\includegraphics[width=0.3\linewidth]{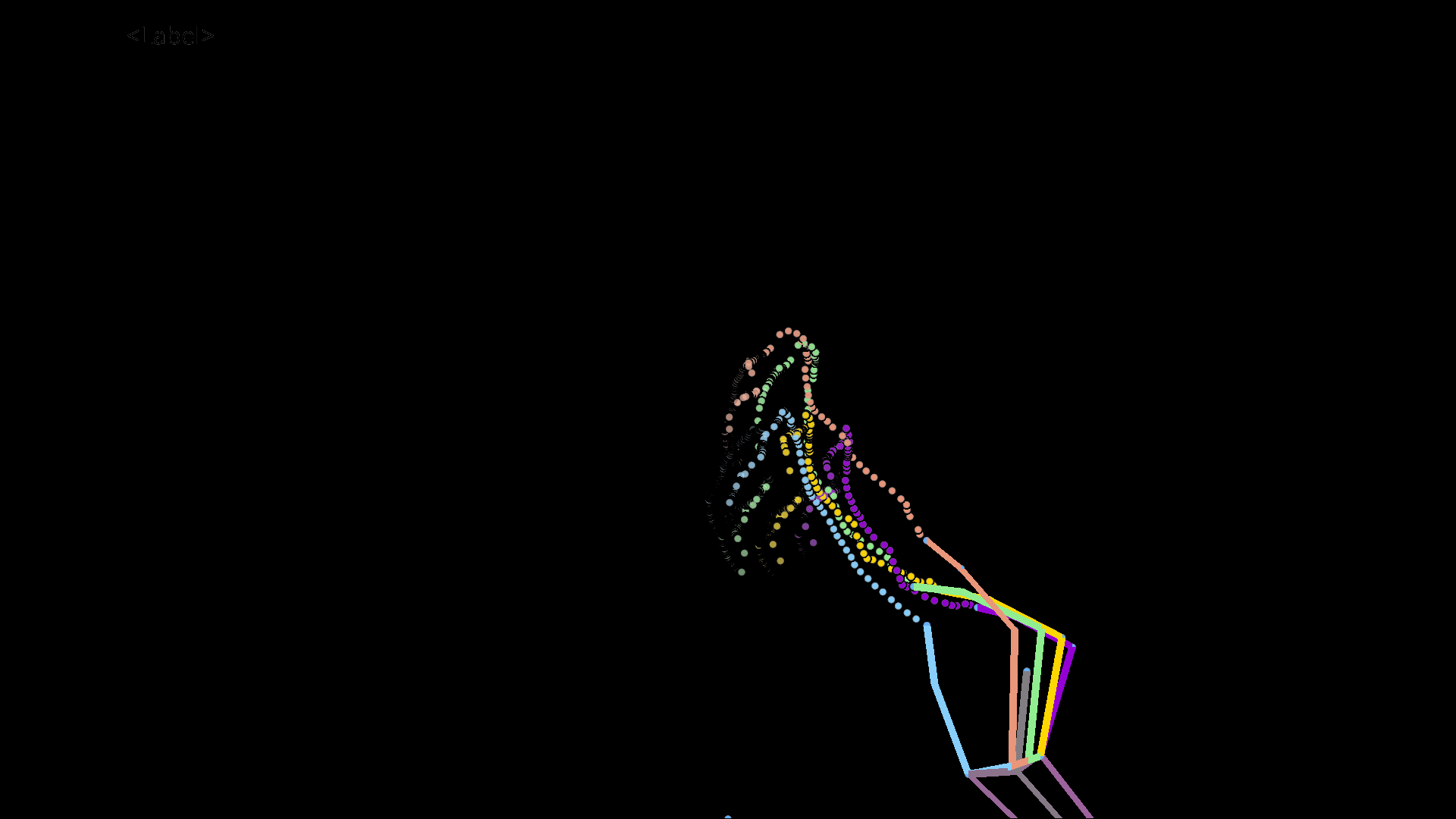}}\quad
	\subfigure[\label{Point8}]{	
		\includegraphics[width=0.3\linewidth]{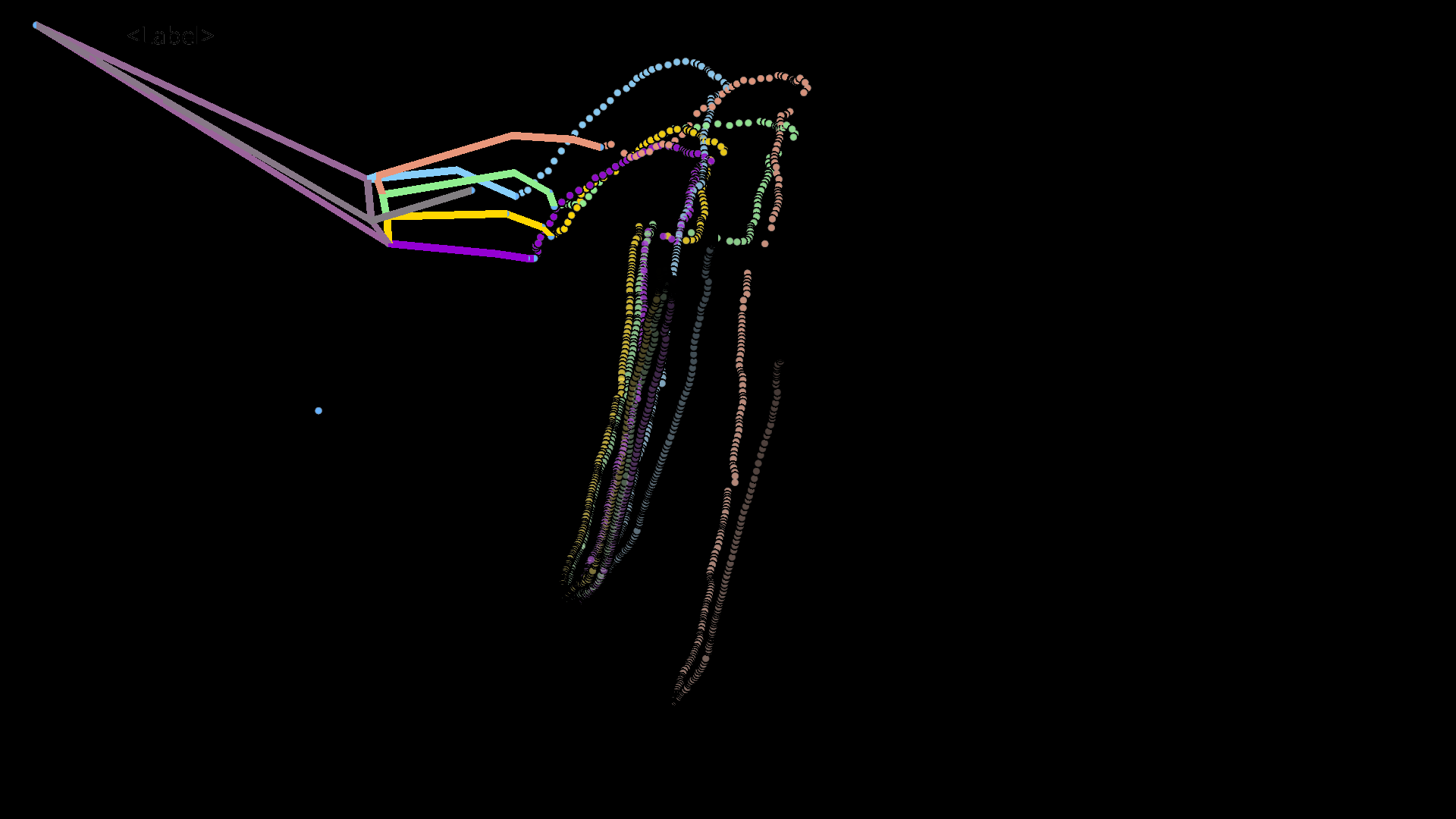}}\quad
	\subfigure[\label{Point9}]{	
		\includegraphics[width=0.3\linewidth]{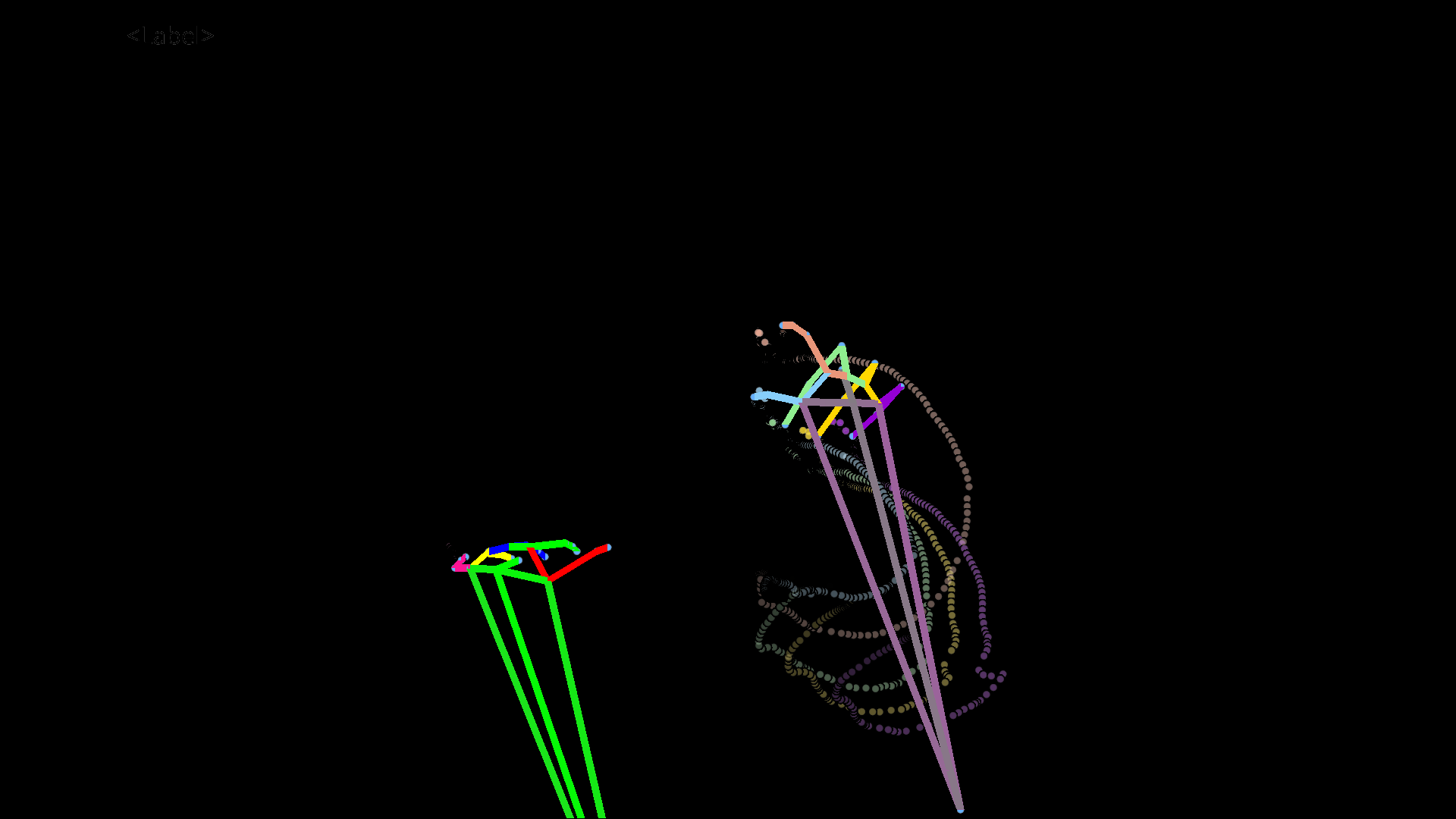}}
	\caption{Examples of \textit{point} patterns present in the training set.}
	\label{fig:pointPattern}
\end{figure}

Using this pattern representation, the accuracy of our method reaches 92.11\%.
This model performs better than the one trained only with top view images, but the improvement is not as significant as we expected. The main reason is that the LMDHG dataset is challenging both in terms of noise, mislabeled samples and for the various semantic interpretations of the gestures which are collected from different users.
\figurename\ \ref{fig:pointPattern} shows different examples of the \textit{Point to} gesture performed by several persons and used to feed the neural network. As can be seen, it is objectively difficult, even for a human being, to be able to distinguish shared characteristics among all the images that univocally indicate that they all belong to the \textit{Point to} class.

\subsection{Evaluation on our new dataset - single view}\label{EvaluationNewDataset}

With the aim of reducing this type of occurrence, we decided to create a new, more balanced dataset\footnote{The images obtained from our new dataset and the ones for the LMDHG dataset are available at this URL: https://github.com/aviogit/dynamic-hand-gesture-classification-datasets}, with more samples per class, and with gestures performed in a more homogeneous and less noisy way. The dataset has around 2000 gesture images sampled every 5 seconds and each class has around 100 samples. The \textit{Rest} class now contains only images of hands that are mostly still. Two further classes have been added: the \textit{Blank} class which contains only traces of gestures that are distant in time (or no gesture at all) and the \textit{Noise} class, which represents all the gestures not belonging to any other class. The dataset is provided both in the form of images and ROS \textit{bags}. The latter can be replayed (in a very similar way to a ``digital tape" of the acquisition) through ROS' \textit{rosbag play} command and this will re-publish all the messages captured during the acquisition (skeleton + depth images) allowing to rerun the pipeline, possibly by changing the processing parameters (e.g. displaying the gestures in a different way or changing the sampling window to improve the real-time acquisition).

Using this new dataset, we then trained a new model, using a 70\%/30\% random split (1348 images for the training set, 577 images for the validation set). The overall accuracy of the model is 98.78\%. We report in \figurename\ \ref{fig:gesturePatternDoubleView} the confusion matrix obtained from this model.
\begin{figure}[htbp]
	\centering
	\includegraphics[height=0.45\textheight]{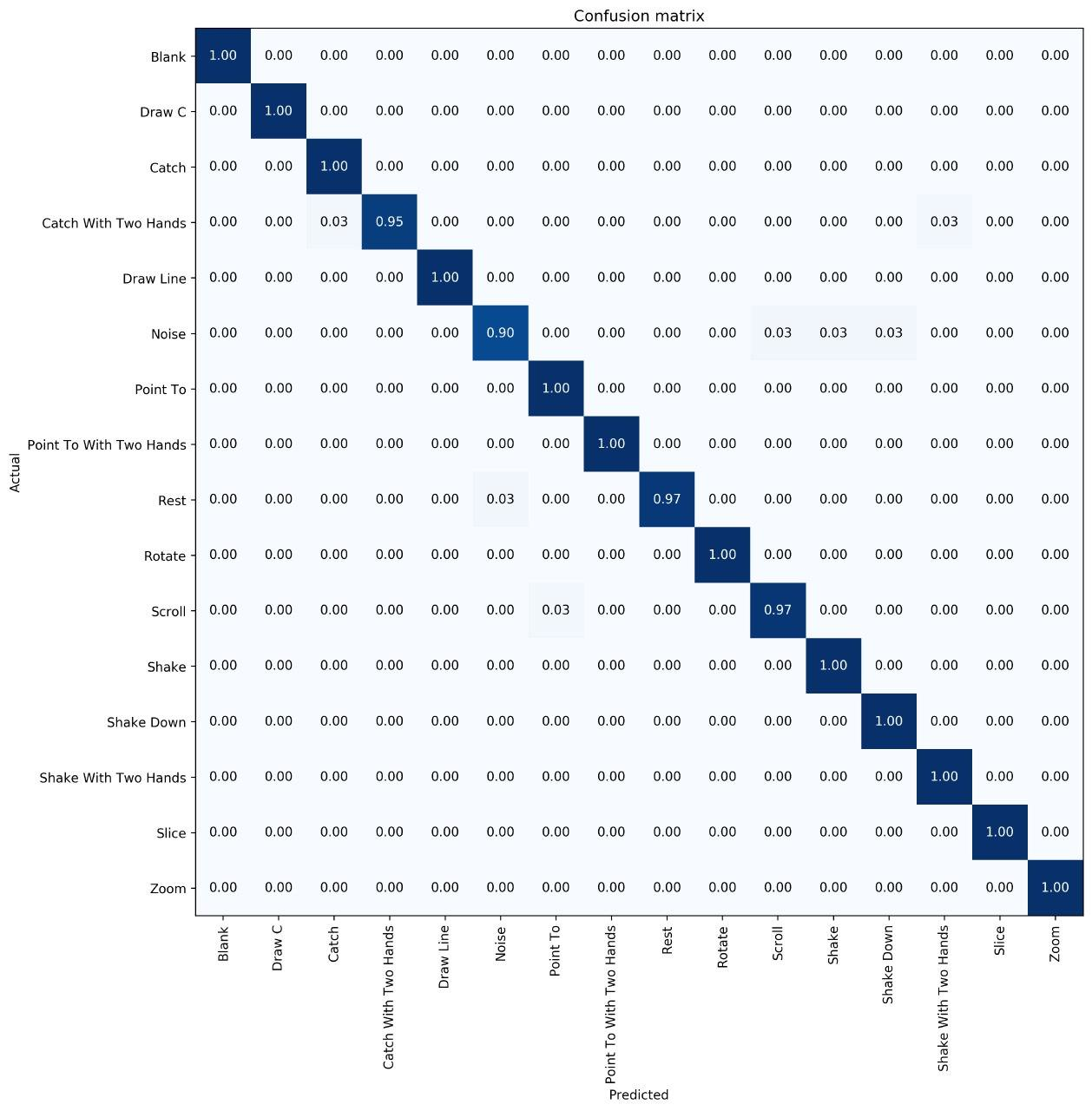}
	\caption{Confusion matrix obtained using the new dataset.}
	\label{fig:ConfMatrixDoubleView}
\end{figure}

\newpage




\subsection{Real-time application}

The real-time acquisition, visualization and classification pipeline has already been used extensively to acquire the new dataset proposed in this paper and for qualitative user tests, again with a sampling window set to 5 seconds. On a PC with an Nvidia GTX 770 GPU, the ResNet-50 model takes a few hundred milliseconds to perform inference on an image produced by the 3D visualizer, thus making the real-time approach usable on practically any machine. However, these tests do not yet have sufficient statistical significance and must therefore be extended to several participants before they can be published. This part will be a subject of future works.



%


\section{Conclusions}\label{Conclusions}

In this paper, we have proposed a visual approach for the recognition of dynamic 3D hand gestures through the use of convolutional neural network models. The pipeline that we propose acquires data (on file or in real-time) from a Leap Motion sensor, it performs a representation in a 3D virtual space from which one or more 2D views are extracted. These images, which condense the temporal information in the form of traces of the fingertips with varying color intensity, are then fed to a CNN model, first in the training phase, then in real-time for the inference phase. The two models trained on the LMDHG dataset achieved an accuracy of above the 91\% and 92\% respectively, while the model trained on the new dataset proposed in this paper reaches an accuracy above the 98\%.

Future work will have the primary objective of enriching the new dataset, both in terms of the number of images, possibly by joining it with the LMDHG dataset after making the appropriate modifications and re-labeling, and in terms of the number of recognized gestures.
In addition, the performance of the real-time pipeline will be validated with a benchmark extended to the largest possible number of users. 

%
%

%
%
\bibliographystyle{splncs04.bst}
\bibliography{bibliography}
\end{document}